\documentclass[11pt, letterpaper]{berkeley}

\usepackage{microtype}
\usepackage{graphicx}
\usepackage{subfigure}
\usepackage{booktabs} 
\usepackage{multirow}
\usepackage{hyperref}
\usepackage{natbib}
\usepackage{wrapfig}
\usepackage{multicol}
\usepackage{fancyvrb}
\usepackage{fvextra}
\usepackage{adjustbox}
\usepackage[hang,flushmargin]{footmisc}

\usepackage[capitalize,noabbrev]{cleveref}

\newcommand{\bigmath}{\textsc{Big-Math}}
\newcommand{\bigmathmc}{\textsc{Big-Math-Reformulated}}

\newcommand{\alon}[1]{\textcolor{orange}{(Alon: #1)}}

\newcommand\blfootnote[1]{%
  \begingroup
  \renewcommand\thefootnote{}\footnote{#1}%
  \addtocounter{footnote}{-1}%
  \endgroup
}

\usepackage{tcolorbox}
\tcbuselibrary{skins}
\newtcolorbox{text_full}[1]{
    enhanced,
    left=4mm,
    right=4mm,
    top=2mm,
    bottom=2mm,
    boxsep=0mm,
    rounded corners,
    title=#1,
    fontupper=\footnotesize\linespread{0.9}\fontfamily{lmr}\selectfont,
    }
\newtcolorbox{text_half}[1]{
    enhanced,
    left=2mm,
    right=2mm,
    top=2mm,
    bottom=2mm,
    boxsep=0mm,
    rounded corners,
    title=#1,
    width=0.475\textwidth,
    fontupper=\footnotesize\linespread{0.9}\fontfamily{lmr}\selectfont,
    }

\title{\bigmath: A Large-Scale, High-Quality Math Dataset for Reinforcement Learning in Language Models}
\author[1]{Alon Albalak}
\author[1]{Duy Phung}
\author[1]{Nathan Lile}
\author[2]{Rafael Rafailov$^\dagger$}
\author[2]{Kanishk Gandhi}
\author[1]{Louis Castricato}
\author[2]{Anikait Singh}
\author[1]{Chase Blagden}
\author[2]{Violet Xiang}
\author[1]{Dakota Mahan}
\author[2]{Nick Haber}

\affil[1]{SynthLabs}
\affil[2]{Stanford University}

\begin{document}

\begin{abstract}
Increasing interest in reasoning models has led math to become a prominent testing ground for algorithmic and methodological improvements.
However, existing open math datasets either contain a small collection of high-quality, human-written problems or a large corpus of machine-generated problems of uncertain quality, forcing researchers to choose between quality and quantity.
In this work, we present \bigmath, a dataset of over 250,000 high-quality math questions with verifiable answers, purposefully made for reinforcement learning (RL).
To create \bigmath, we rigorously filter, clean, and curate openly available datasets, extracting questions that satisfy our three desiderata: (1) problems with uniquely verifiable solutions, (2) problems that are open-ended, (3) and problems with a closed-form solution.
To ensure the quality of \bigmath, we manually verify each step in our filtering process and iteratively improve our filters over multiple rounds.
Based on the findings from our filtering process, we introduce 47,000 new questions with verified answers, \bigmathmc: closed-ended questions (i.e.\ multiple choice questions) that have been reformulated as open-ended questions through a systematic reformulation algorithm.
Compared to the most commonly used existing open-source datasets for math reasoning, GSM8k and MATH, \bigmath~is an order of magnitude larger (250,000 questions vs. 8,000 questions in GSM8k and 12,000 in MATH), while our rigorous filtering ensures that we maintain the questions most suitable for RL.
We also provide a rigorous analysis of the dataset, finding that \bigmath~contains a high degree of diversity across problem domains, and incorporates a wide range of problem difficulties, enabling a wide range of downstream uses for models of varying capabilities and training requirements.
In conclusion, this work presents our new dataset, \bigmath, the largest open dataset of math problems suitable for RL training. By bridging the gap between data quality and quantity, \bigmath~establish a robust foundation for advancing reasoning in LLMs. \bigmath~ and \bigmathmc~are available at \url{https://huggingface.co/datasets/SynthLabsAI/Big-Math-RL-Verified}.

\end{abstract}

\maketitle

\blfootnote{
Filtering and reformulation code available at \url{https://github.com/SynthLabsAI/Big-Math}.\\
Prompts for all filters and reformulation can be found in Appendix~\ref{app:prompts}.\\
$^\dagger$Work done while at SynthLabs.
}







\section{Introduction}


In the past few years, mathematics has emerged as a critical testing ground for the development and evaluation of advanced reasoning techniques used in Large Language Models (LLMs)~\citep[inter alia]{cobbe2021trainingverifierssolvemath, wei2023chainofthoughtpromptingelicitsreasoning, schaeffer2023emergentabilitieslargelanguage, mirzadeh2024gsmsymbolicunderstandinglimitationsmathematical, openai2024openaio1card, deepseek2024r1}.
Following the release of strong reasoning models, such as OpenAI's o1~\citep{openai2024openaio1card} and DeepSeek's R1~\citep{deepseek2024r1}, a plethora of supervised fine-tuning (SFT) datasets have been released with the aim to distill reasoning capabilities into other models, or to bootstrap a model prior to RL~\citep{liu2024acemathadvancingfrontiermath,li2025llmseasilylearnreason,servicenow_r1distill,bespoke_stratos,muennighoff2025s1simpletesttimescaling}.
However, knowledge distillation has its limitations. For example, research has shown that while SFT distillation enables models to memorize reasoning patterns and solutions, the resultant models do not generalize well~\citep{chu2025sftmemorizesrlgeneralizes}. On the other hand, RL training algorithms have been shown to yield models with better generalization to novel problems by emphasizing the exploration and refinement of reasoning strategies.
Furthermore,~\citep{wang2025examiningfalsepositivesinference} find that current reasoning models (across model sizes and families) frequently achieve the correct answer with incorrect reasoning steps, suggesting that distilled reasoning models will also suffer from incorrect reasoning. Additionally,~\citet{openai2025competitiveprogramminglargereasoning} show that scaling RL training is very effective in improving reasoning and other capabilities.
Therefore, \textit{while distillation through SFT datasets is clearly a helpful method for improving model performance on some evaluations, it does not address the greater needs at the forefront of reasoning capabilities which can be addressed by RL: learning how to reason}~\citep{xiang20252reasoningllmslearning, yeo2025demystifyinglongchainofthoughtreasoning, kim2025metastabledynamicschainofthoughtreasoning}.

A significant bottleneck in RL-based reasoning research is the lack of high-quality datasets tailored for reinforcement learning.
RL training methods all assume access to a dataset with verifiable answers \citep{deepseek2024r1, kimiteam2025kimik15scalingreinforcement}; either to train a reward model~\citep{lightman2023letsverifystepstep, havrilla2024glorewhenwhereimprove, zhang2024generativeverifiersrewardmodeling, mahan2024generativerewardmodels} or to directly evaluate the correctness of a generated solution and, as has been demonstrated through the course of deep learning's existence, scaling up data quantity and quality is a crucial step towards success~\citep{havrilla2024surveyingeffectsqualitydiversity}.
However, \textit{a central issue with the existing math datasets is that they are either (1) human written, but limited in quantity, or (2) machine generated and large, but of unknown quality, forcing researchers to choose datasets with either quality or quantity.}
For example, prior works have mostly used the GSM8k~\citep{cobbe2021trainingverifierssolvemath} and MATH~\citep{hendrycks2021measuringmathematicalproblemsolving} datasets. While they contain human-written questions and answers, both datasets are quite limited in quantity, with 8,000 and 12,000 problems, respectively.
On the other hand, large-scale datasets, such as NuminaMath~\citep{numina_math_datasets}, exhibit quality issues, including many duplicate problems and incomplete solutions, hindering their utility in RL training.
Furthermore, many existing math datasets contain problems which are not well-suited for training a reasoning model with RL. For instance, they sometimes have a high proportion of multiple choice questions. While multiple choice questions may be useful for SFT, they are less effective for RL-based training. Recent works~\citep{xiang20252reasoningllmslearning}, suggest that the goal of RL for reasoning is not just to train the model to answer correctly, but \textit{more importantly}, to reason correctly. With this in mind, questions in multiple choice formats are problematic for RL; even though the correct answers may be difficult to deduce, the model can simply guess the correct answer without performing the correct reasoning.
Problems like these highlight the biggest issues with curating a dataset for RL.

To address these challenges, we present \bigmath, a dataset of over 250,000 high-quality math problems and solutions, curated with three core desiderata:
\begin{enumerate}
    \item \textbf{Uniquely verifiable solutions:} problems must admit a single correct answer that can be reliably verified;
    \item \textbf{Open-ended problem formulations:} problems that cannot be easily solved by guessing (as might occur in multiple choice formats) and instead require nontrivial reasoning steps; and
    \item \textbf{Closed-form solutions:} answers must be expressible in a closed form (e.g., a scalar or formula, not a proof), thereby enabling automated evaluation.
\end{enumerate}
With these desiderata in mind, we design and develop a process of cleaning and curating datasets by applying a strict set of filters. To ensure the quality of our filtration process, we apply a human-in-the-loop algorithm, iteratively improving each filter through multiple rounds of manual verification until the filter achieves suitable levels of precision and recall (minimum of 90\%). Then, we apply our filtration process to three openly available datasets, extracting a high-quality and large-scale subset of over 200,000 uniquely verifiable, open-ended problems with closed-form solutions.
Additionally, we introduce \bigmathmc, a novel subset of 47,000 problems derived by reformulating multiple-choice questions into an open-ended format, while preserving their integrity and complexity and aligning with RL training requirements.
In combination, our filtered datasets and reformulated problems create~\bigmath, a dataset of over 250,000 questions with extracted answers, specifically designed for RL training.

In summary, this work provides:
\begin{itemize}
    \item \bigmath, the largest open source dataset of math problems, designed specifically for RL, with over 250,000 problems.
    \item A rigorous filtering and cleaning pipeline, involving humans-in-the-loop, and achieving over 90\% precision and recall.
    \item \bigmathmc, a dataset of 47,000 novel problems, reformulated from existing multiple choice problems, and the full reformulation pipeline.
    \item Empirical analyses highlighting the diversity, difficulty, and suitability of~\bigmath~for advancing reasoning in LLMs.
\end{itemize}

\begin{figure}[t]
    \centering
\begin{minipage}{.475\textwidth}
    \centering
    \includegraphics[width=\textwidth]{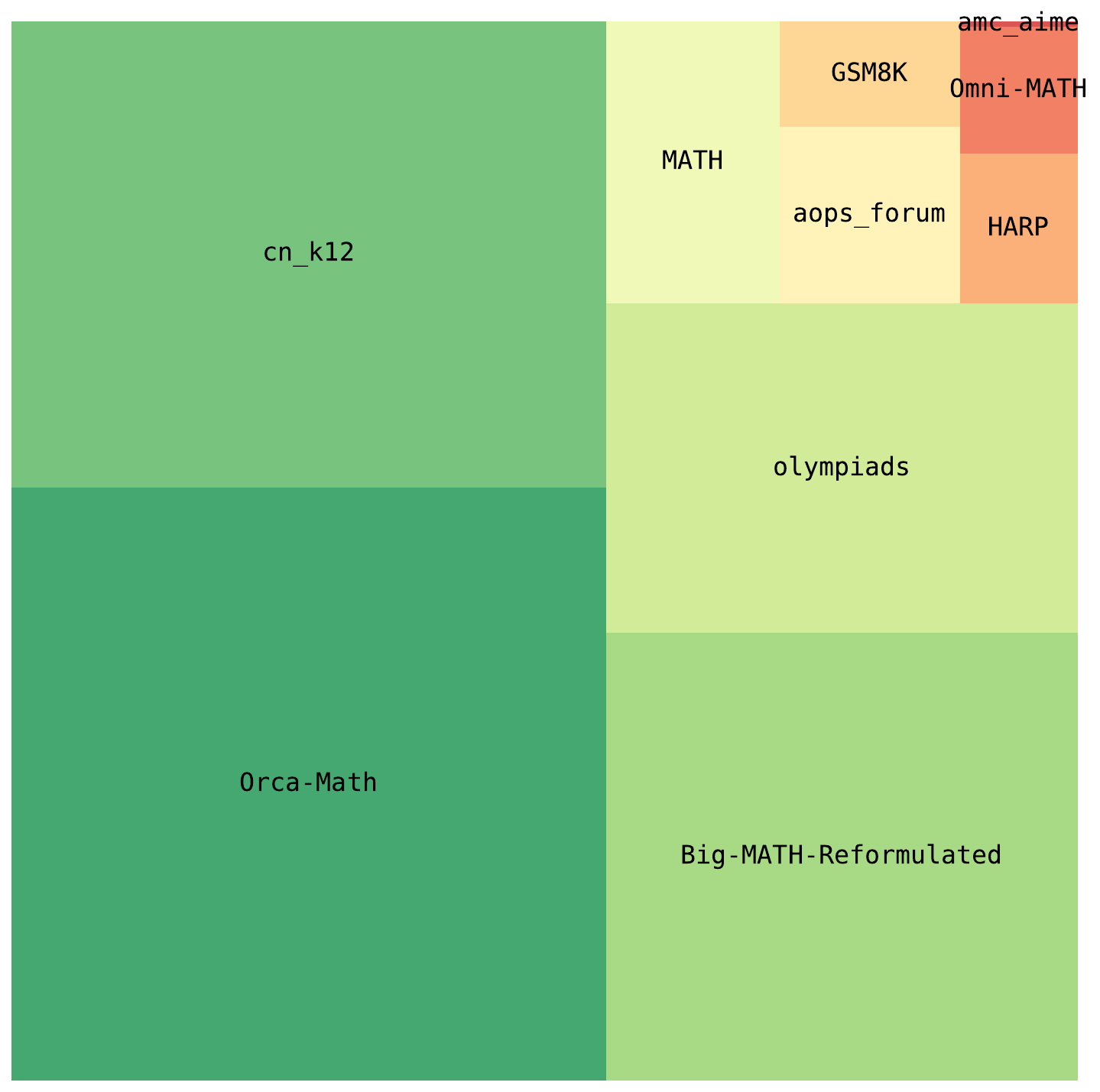}
\end{minipage}%
\begin{minipage}{.5\textwidth}
    \centering
    \includegraphics[width=\textwidth]{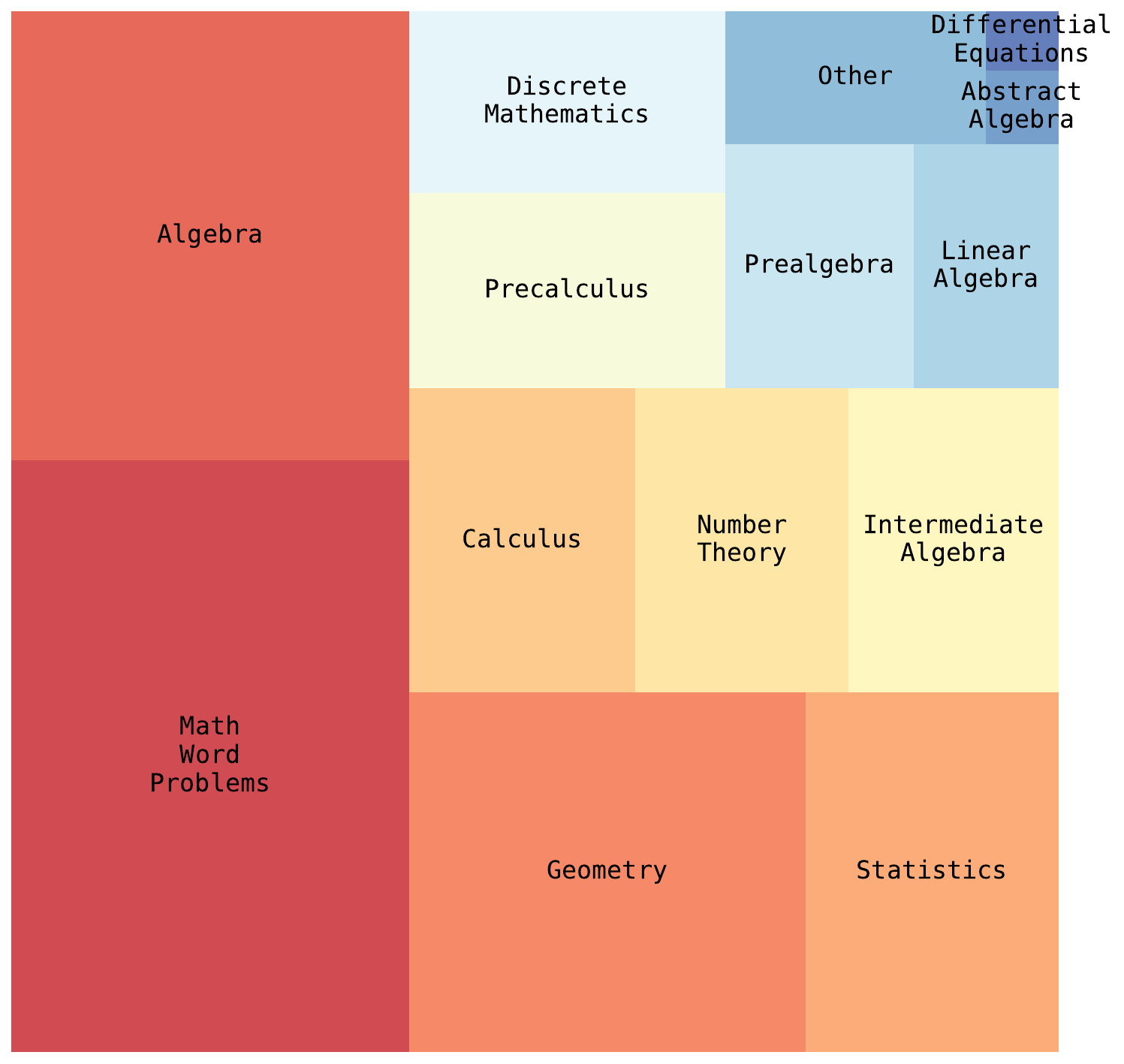}
\end{minipage}
\caption{The sources (left) and domains (right) of the \bigmath~dataset, scaled by size.}
\label{fig:treemaps}
\end{figure}

\begin{table*}
\centering
\begin{tabular}{|l|r|r|}
\toprule
 \textbf{Data Source}& \textbf{Original} & \textbf{\bigmath} \\
\midrule
Orca-Math~\citep{mitra2024orcamathunlockingpotentialslms} & 153,314 & 83,215 \\
cn\_k12~\citep{numina_math_datasets} & 276,554 & 63,609 \\
olympiads~\citep{numina_math_datasets} & 150,563 & 33,485 \\
MATH~\citep{hendrycks2021measuringmathematicalproblemsolving} & 12,000 & 9,257 \\
aops\_forum~\citep{numina_math_datasets} & 30,192 & 5,740 \\
GSM8k~\citep{cobbe2021trainingverifierssolvemath} & 7,473 & 3,254 \\
HARP~\citep{yue2024harp} & 4,780 & 2,996 \\
Omni-MATH~\citep{gao2024omnimathuniversalolympiadlevel} & 4,428 & 2,478\\
amc\_aime~\citep{numina_math_datasets} & 4,070 & 78 \\
\bigmathmc~(Section~\ref{sec:bigmathmc}) &
-- & 47,010 \\
\hline
\textbf{Total} & 643,374 & 251,122 \\
\bottomrule
\end{tabular}
\caption{\textbf{Comparison of problems by data source} with the original quantity and the quantity included in \bigmath. \bigmathmc~is a new problem set introduced in this work and does not have an ``original'' size.}
\label{tab:data_sources}
\end{table*}

\section{\bigmath}

In this section, we describe the technical details of our data collection, cleaning, filtering, and reformulation processes.

\subsection{Dataset Collection}





We considered a number of openly available math datasets and selected 3 well established mathematical problem datasets that are commonly used in recent literature: HARP~\citep{yue2024harp}, Omni-MATH~\citep{gao2024omnimathuniversalolympiadlevel}, and NuminaMath~\citep{numina_math_datasets}.
The goal of this work is to find a large set of high-quality math problems that satisfy our three desiderata. Human-written problems are generally associated as being high-quality, and the datasets that we select provide a large quantity of human-written problems to start with ($> 75\%$). Additionally, we choose to incorporate synthetically generated data, but limit ourselves to a single, well established source
Table~\ref{tab:data_sources} shows a breakdown of the sources, including the original quantity of data and the amount after filtering.

\paragraph{HARP} First, we include the Human Annotated Reasoning Problems (HARP) dataset~\citep{yue2024harp}, which contains 4,780 short answer problems collected from U.S. national math competitions. We specifically take only the ``short answer'' subset, which should not contain any questions requiring proofs, or multiple choice questions.

\paragraph{Omni-MATH} Next, we incorporate the Omni-MATH dataset~\citep{gao2024omnimathuniversalolympiadlevel}, contributing almost 4,500 olympiad-level problems curated from 39 different competition websites. Notably, this dataset employs professional annotators and verifiers to maintain solution-answer quality.

\paragraph{NuminaMath} Finally, we use the NuminaMath~\citep{numina_math_datasets} dataset, which is composed of roughly 860,000 problems divided into 9 sources (6 of which we retain).
First, there are three sources that we do not use from Numina: the synthetic\_math, synthetic\_amc, and the MATH subsets. We choose not to include either of the synthetic data subsets as they was never evaluated independently from the remainder of the dataset, so we cannot know their quality. Additionally, while NuminaMath uses the original split of the MATH dataset (7,000 training problems, 5,500 test problems), we choose to use version with 12,000 training problems and 500 test problems, as proposed by~\citep{lightman2023let}.
The largest subset from Numina is the cn\_k12 subset of data, composed of $\sim275,000$ math problems scraped from chinese math exams. The next largest subset from NuminaMath is the Orca-Math~\citep{mitra2024orcamathunlockingpotentialslms} subset ($\sim150,000$ problems), which is a synthetically generated dataset of grade school math problems. Although this data was synthetically generated,~\citet{mitra2024orcamathunlockingpotentialslms} demonstrate that as the lone source of SFT training data, this subset leads to impressive math performance, even for small language models. Next, we include the olympiads subset, consisting of 150,000 problems collected from online sources of international-level math competitions. Next,~\citet{numina_math_datasets} scrape the Art of Problem Solving forum for additional math competition problems, selecting to keep only problems with high quantities of LaTeX and at least one \texttt{\textbackslash boxed} or $\blacksquare$ symbol, leading to an additional 30,000 problems. Additionally, we use the GSM8k~\citep{cobbe2021trainingverifierssolvemath} subset, providing another $\sim8,000$ problems. Finally, we also include the amc\_aime subset, consisting of 4,000 more math competition problems, with solutions scraped from online sources.

\subsection{Dataset Cleaning and Filtering}
The collection of all the above datasets leads us to a combined dataset of over 640,000 problems. However, this dataset likely has many duplicated questions, undesirable content, and data that does not satisfy our desiderata. To achieve a dataset of the highest quality, and appropriate for RL training, we next clean and filter the data from each source using a combination of bespoke and common strategies~\citep{albalak2024surveydataselectionlanguage, soldaini2024dolmaopencorpustrillion, li2024datacomplmsearchgenerationtraining}.
We ensure the quality of our filters by applying human-in-the-loop methodology, iteratively refining the filters through human verification and annotation of positive and negative examples. By the end of the iterative process our filters achieve over 90\% F1 score, oftentimes reaching much higher than 90\%. After filtering, the data should contain only problems that closely follow our three desired properties: open-ended, verifiable, closed-form problem-solution pairs.

\subsubsection{Source-specific Filtering and Cleaning}
The first step in our cleaning and filtering process is to observe a sampling of data from each dataset, and to design bespoke filters to be utilized on each source separately based on their unique idiosyncrasies. 
\paragraph{HARP}
For the HARP dataset~\citep{yue2024harp}, we find many problems that contain figures in the Asymptote\footnote{\url{https://asymptote.sourceforge.io/}} vector graphics language, which we identify by string matching for ``[asy]''. We err on the side of caution and remove 625 such problems (13\% of the dataset), assuming that the model would need to see the rendered image to solve the question.
\paragraph{Omni-MATH}
When exploring Omni-MATH~\citep{gao2024omnimathuniversalolympiadlevel}, we found a number of problems containing author attributions (e.g.\ a person's name in parenthesis, or in the format ``[i] Name [/i]'') and manually revise each problem to remove the attribution (45 in total). Next, we found that some problems contain information about their scores in the competition and remove this text from the problem. Specifically, we found the phrase ``If the correct answer is \texttt{X} and your answer is \texttt{Y}'' in a number of problems. Additionally, we found and removed 2 problems with the following solution: ``The problem provided does not contain a solution. Therefore, no final answer can be extracted.'', likely a parsing error when extracting problem-solution pairs from online sources.
\paragraph{NuminaMath}
Within NuminaMath~\citep{numina_math_datasets}, we find some unique characteristics of each subset. First, because some of the subsets are quite large, we deduplicate problems within each subset with a MinHashLSH filter. We use 128 hashing functions, and through a few rounds of experiments, empirically determine that a similarity threshold of 0.6 or 0.7 (depending on the subset) is the strictest threshold we can set without the filter quality degrading.
Next, the NuminaMath dataset does not explicitly contain answers to each problem, so we extract answers by searching for boxed solutions (``\textbackslash boxed\{\}'' in LaTeX). Any problem whose solution does not contain exactly 1 boxed answer is filtered out.
Finally, for the aops\_forum subset, we find 2535 problems containing unnecessary information, including problem attribution (e.g.\ ``proposed by''), year of submission, point scoring (e.g\ ``(1 point)'') and remove these strings with regular expressions.

\noindent At the end of source-specific filtering, we find that we have 463,426 remaining problems. However, these problems are not yet fit for use, as this data still contains problems not suitable for RL training.

\begin{table}[t]
\centering
\small
\begin{tabular}{|l|rrrrrr|r|}
\toprule
& \textbf{Multiple} & \multirow{2}{*}{\textbf{True/False}} & \multirow{2}{*}{\textbf{Yes/No}} & \multirow{2}{*}{\textbf{Multi-Part}} & \multirow{2}{*}{\textbf{Proof}} & \textbf{Semantic} & \textbf{Total} \\
\textbf{Subset} & \textbf{Choice} & & & & & \textbf{Duplicates} & \textbf{Matching} \\
\midrule
Orca-Math & 1,277 & 5 & 12 & 18,375 & 5 & 8,286 & 26,487 \\
cn\_k12 & 96,804 & 420 & 384 & 38,625 & 5,233 & 2,098 & 123,809 \\
olympiads & 12,688 & 311 & 4,615 & 22,116 & 30,357 & 1,362 & 53,102 \\
MATH & 1,124 & 0 & 3 & 2,885 & 31 & 7,694 & 9,714 \\
aops\_forum & 1,665 & 85 & 126 & 3,162 & 4,608 & 262 & 7,853 \\
GSM8k & 46 & 0 & 0 & 792 & 0 & 7,664 & 7,999 \\
HARP & 303 & 2 & 7 & 288 & 2 & 155 & 695 \\
Omni-MATH & 149 & 126 & 206 & 467 & 207 & 378 & 1,182 \\
amc\_aime & 2,962 & 0 & 22 & 1,274 & 66 & 53 & 3,066 \\
\midrule
\textbf{Total} & \multirow{2}{*}{117,018} & \multirow{2}{*}{949} & \multirow{2}{*}{5,375} & \multirow{2}{*}{87,984} & \multirow{2}{*}{40,509} & \multirow{2}{*}{27,952} & \multirow{2}{*}{233,907} \\
\textbf{Matching} & & & & & & & \\
\bottomrule
\end{tabular}
\caption{\textbf{The quantity of data matched by each filter.} Some data matches multiple filters, so column totals are a summation, but row totals are not. We see some interesting trends here: the cn\_k12 subset contains a significant amount more multiple choice questions than any other subset, olympiads contains many more proof questions, but also yes/no questions than other subsets. Interestingly, but unsurprisingly, we also see that nearly the entire MATH and GSM8k datasets contain semantic duplicates with other subsets.}
\label{tab:filter_result}
\end{table}

\subsubsection{Source-agnostic Filtering}
\label{sec:source_agnostic_filters}
After running each of the described filters over the individual subsets, we perform 11 filtering operations across the full collection. These filters are specially designed to convert the raw dataset of question-answer pairs into a dataset that is suitable for training a math reasoning model with RL.
Just as for the source-specific filters, we also manually verify the results of each source-agnostic filter, and iteratively improve their performance until they achieve suitable results (where appropriate). The high-level results of our filtering process are found in Table~\ref{tab:filter_result}, with more detailed results in Appendix~\ref{app:filter_result_expanded}. Prompts for model-based filters are found in Appendix~\ref{app:filter_prompts}.




\paragraph{Deduplication and Decontamination}
To prevent a model from unknowingly being trained on the same problem too frequently (and risk overfitting), we need to remove duplicates.
To handle this, we run a very simple and efficient deduplication step, calculating duplicate problem by string matching (not including whitespace), removing all but one copy of the duplicated problem.
Next, we also want to ensure that the final dataset has a diverse set of problems, without focusing on any specific problem types. To do so, we remove semantic duplicates (problems with similar meaning, e.g.\ the same problem with numbers changed) with the SemDeDup algorithm~\citep{abbas2023semdedupdataefficientlearningwebscale}. To embed the problems, we use the model at \href{https://huggingface.co/sentence-transformers/all-MiniLM-L6-v2}{sentence-transformers/all-MiniLM-L6-v2} and remove problems with a cosine similarity over 0.5. For semantic deduplication, we tested thresholds between 0.95 and 0.2, finding that even the strictest thresholds can cluster dissimilar problems (likely caused by short problems). We want to ensure that no duplicated data passes through, so we purposefully select a threshold where our manual verification determines that no duplicates remain.
Finally, the MATH~\citep{hendrycks2021measuringmathematicalproblemsolving} and Omni-MATH~\citep{gao2024omnimathuniversalolympiadlevel} test sets are prime candidates for evaluating a model trained on our dataset, so we need to ensure that the problems in those test sets do not exist in our training set.
To decontaminate our dataset with any of the 500 MATH test set problems and the 500 Omni-MATH test set problems we find and remove all contaminated data by a string matching algorithm on the problems, the same as for deduplication.
We were surprised to find minimal duplication ($<1\%$), removing only 4,229 problems, with the majority of duplicates being between the Orca-Math (3,916), GSM8k (3,875), and cn\_k12 (405) datasets.

\paragraph{Ensuring that problems are solvable}
For a number of reasons, there are problems in existing math datasets which are not solvable.\\
\indent \textbf{(Language Filter)}
First, in this work, we focus on English-only models and require English math problems. So, we use a FastText language identifier~\citep{joulin2016bag, joulin2016fasttext, grave2018learning} and remove any problems where English is not the primary language (only 101 non-English problems were detected).\footnote{We use the model at \url{https://huggingface.co/facebook/fasttext-language-identification}.} Through our iterative improvement process, we found that it was important to remove LaTeX, along with most special characters (e.g.\ numbers, math symbols, ``()[]\{\}!@\#\$\%\^{}\&\*'', etc.) in order to achieve a high level of accuracy with the language detection model. Additionally, problems which were very short ($<10$ characters) were often classified as non-English (even if they were entirely numbers), so we simply include all problems with fewer than 10 non-LaTeX, non-special characters.\\
\indent \textbf{(Hyperlink Detection)}
Next, we remove problems containing a hyperlink using a simple regular expression, as the existence of hyperlinks suggests that a model may not have the full resources required to solve the problem (e.g.\ hyperlinks that point to a website containing a theorem). While this likely removes problems that are solvable (hyperlinks may also link to the website source of the problem), we prefer to err on the side of caution and remove all problems containing hyperlinks. This over-filtering can be addressed in the future through more complex filters.\\
\indent \textbf{(Model Solve Rate)}
Finally, while it is not feasible to manually ensure the correctness of each problem-answer pair, we develop a heuristic for correctness using language models. For each problem, we generate 64 solutions from Llama-3.1-8B ($\sim 30,000,000$ rollouts) and 5-8 solutions from Llama-3.1-405B ($\sim 1,100,000$ rollouts, generated on a pre-filtered subset)~\citep{dubey2024llama3herdmodels}. If either model answers the question with the ground truth answer, then we determine that the question-answer pair may be valid. We do not apply this filter to HARP, Omni-Math, MATH, or GSM8k as these datasets include pre-parsed answers.\\
\indent Of course, this method does not guarantee that the given answer is correct, as it is possible that the answers fall under a commonly made mistake, or that the models have seen the data during pre- or post-training. Furthermore, this filter does not guarantee that removed data has an incorrect answer, as it is very likely that the models we use cannot solve the most difficult math problems. One method for improving this filter would be to use a stronger, math-specific model.

\paragraph{Ensuring that problems are open-ended}
An important aspect of reinforcement learning is that the training signal should appropriately attribute the good actions with high rewards and poor actions with low rewards. Therefore, problems with multiple choice answers pose a problem: the model can inadvertently respond with the correct answer option (generally between a 25-50\% chance of guessing correctly) without providing the correct intermediate reasoning steps, leading to a poor learning signal. For this reason, we choose to remove any problems that are multiple choice, True/False, and Yes/No.
To detect and remove all three types of questions, we develop both a regular expression-based filter, and a model-based filter.\\
\indent \textbf{(Regular Expression Filters)}
For multiple choice questions, we use a simple regular expression filter that searches for either alphabetic options (A, B, C, D) or numerical options (1, 2, 3, 4), occurring in order. To ensure that we do not incidentally remove questions referring to shapes (e.g.\ ``rectangle ABCD...'') or numbers (e.g.\ 1234), we first remove those strings from the question, prior to the regular expression search.
Next, for the True/False questions, we search for either ``true'' or ``false'' in the answer or, when available, in the final line of the solution.
Then, for Yes/No questions, we perform the same check as True/False questions, searching in the answer or solution for exact phrases. Additionally, we search the final line of the question for specific phrases that imply a Yes/No question: ``is'', ``are'', ``do'', ``does'', and ``can''.
While these regular expression filters are guaranteed to remove all data of this exact form, they are not flexible enough to catch many exceptions.\\
\indent \textbf{(Model-based Filters)}
Therefore, for each question type, we design a model-based filter by iteratively developing a prompt to use with Llama-3.1-70B~\citep{dubey2024llama3herdmodels}. For example, to develop the prompt for multiple choice questions, we first manually find examples of multiple choice problems and open-ended problems and include these as in-context examples. Next, we run our filter over the dataset and inspect 100 problems classified as positive examples (multiple choice) and 100 problems classified as negative examples (open-ended). We iteratively add the difficult incorrectly classified problems into the prompt, mostly selecting problems following a previously unseen pattern. For these filters we prioritize a high recall, ensuring that we remove as much undesirable data as possible. In this iterative process of filter development, we continue until achieving over 98\% recall, requiring between 5-8 rounds of manual verification.

\paragraph{Ensuring that problems are uniquely verifiable}
A critical aspect of training reasoning models with reinforcement learning is the existence of verifiable answers, generally in the form of a ground truth to be compared against the model response.\\
\indent \textbf{(Answer Filter)}
As a simple first step, we remove all examples where the final answer did not previously exist, or could not be extracted from the solution (e.g.\ if there was no ``\textbackslash boxed\{\}'' element in the solution).\\
\indent \textbf{(Multi-part Question Filter)}
Next, we find a large proportion of questions with multiple parts requiring multiple corresponding answers. There are surely methods with which to handle partial correctness in multi-part questions, but this is still an open research question requiring further study, so we leave these problems for future, more difficult, versions of the dataset.
To develop our multi-part question filter, we follow a similar pattern to the multiple choice questions filter: a dual-filter approach including a regular expression filter and a model-based filter. For the regular expressions, we iteratively improve the filter through manual search of the data and improvements to the regular expressions. The final version of our filter searches for commonly found signals: ordered roman numerals (e.g.\ \MakeUppercase{\romannumeral 1, \romannumeral 2}), multiple numbered parts in parentheses (e.g.\ (1) ... (2)), multiple numbered parts with a period (e.g.\ 1. ... 2.), as well as numbered special characters (e.g.\ \textcircled{1} ... \textcircled{2}). For the model-based filter, we use the same iterative process as for the multiple choice question, using a Llama-3.1-70B~\citep{dubey2024llama3herdmodels}, inspecting 100 examples of positively and negatively classified examples at each iteration, ending only once we achieve over 98\% recall.\\
\indent \textbf{(Proof Filter)}
While proofs can be verified, there can be many correct variations, and how to quickly verify proofs written in natural language is unclear at the moment (other than converting to a theorem proving language, which can incur additional parsing errors). Therefore we also elect to remove proofs from our dataset, leaving them for a future version which is more difficult.
We develop our proof filter with the same dual-method approach previously discussed. The regular expression filter simply searches for either of the following phrases in the problem: ``prove that'' or ``a proof''. We again implement the model-based filter through an iterative improvement process with Llama-3.1-70B, inspecting 100 examples of predicted positive and negative examples, completing the process only once 98\% recall is achieved.


\begin{figure}
    \centering
    \includegraphics[width=0.95\linewidth]{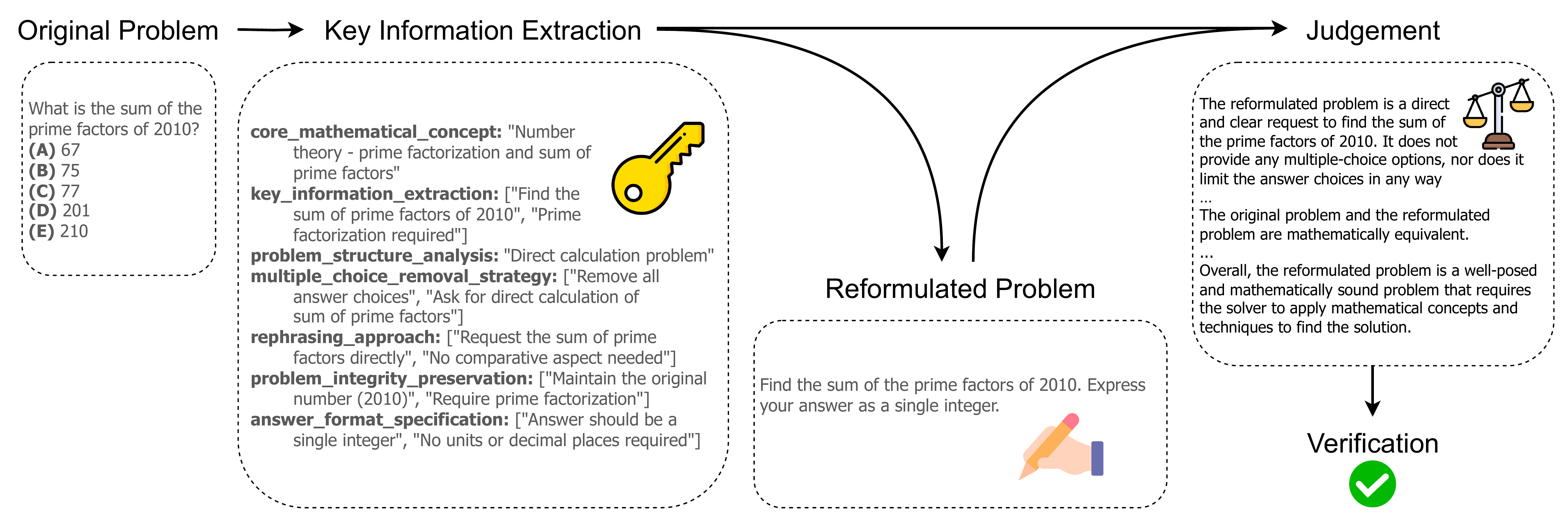}
    \caption{\textbf{Our reformulation strategy} first extracts key information, then reformulates the problem as open-ended, followed by a judgement of the reformulation and a final verification. The full process is detailed in Section~\ref{sec:reformulation_strategy}, with successful and failed examples of reformulation in Appendix~\ref{app:reformulation_examples}.}
    \label{fig:reformulation}
\end{figure}

\subsection{\bigmathmc}
\label{sec:bigmathmc}
Based on findings from the previous sections we introduce a new subset of 47,000 questions and answers, \bigmathmc, which we describe here.

During the development of our filters, we found a staggering number of multiple choice questions ($> 117,000$), which are removed during the filtering process for \bigmath. In particular, we are concerned with the loss of significant quantities of data from human-written and high-quality sources: olympiads, amc\_aime, and aops\_forum.
The inherent structure of multiple choice questions presents a challenge for RL algorithms.
Specifically, the multiple-choice format increases the probability of answering correctly regardless of the correctness of a reasoning chain. This can lead the algorithm to assign high rewards to incorrect reasoning, reducing the utility of such questions in an RL environment.
To address this issue, we propose a novel approach to reintroduce these valuable questions by reformulating the multiple choice questions into open-ended questions with a series of carefully devised and detailed steps enacted by large language models.
Our reformulation process is developed to be systematic and rigorous, in order to ensure that the integrity and complexity of the original questions are maintained, while making them amenable for RL. We include examples of a successful reformulation and a failed reformulation in Figure~\ref{fig:reformulation_examples}.

\subsubsection{Reformulation strategy}
\label{sec:reformulation_strategy}
\begin{wraptable}{r}{0.35\textwidth}
\vspace{-14pt}
    \centering
    \begin{tabular}{|l|r|}
        \toprule
        \multirow{2}{*}{\textbf{Source}} & \textbf{Number of} \\
        & \textbf{Problems} \\
        \midrule
        cn\_k12 & 35,211 \\
        olympiads & 9,019 \\
        amc\_aime & 1,878 \\
        aops\_forum & 853 \\
        MATH & 48 \\
        Orca-Math & 1 \\
        \bottomrule
    \end{tabular}
    \caption{\textbf{Composition of problems in \bigmathmc}, reformulated from problems in the source column.}
\label{tab:bigmathmc}
\end{wraptable}
We develop a 4-step process to reformulate multiple choice problems as open-ended problems. All prompts can be found in Appendix~\ref{app:reformulation_prompts}, and we use Llama-3.1-405B~\citep{dubey2024llama3herdmodels} to create \bigmathmc.\\
\indent \textbf{(Key Information Extraction)} We begin the process by identifying a few core pieces of information about each question. We determine whether the question is in the multiple choice format, extract the core mathematical concepts (e.g.\ geometry), and identify key problem details (e.g.\ distances, goal of the question). In the first step, we also ask the language model to develop a strategy to convert the question into an open-ended format, as well as strategies for rephrasing and ensuring that the integrity of the original problem is maintained. Finally, we extract a plan for what the format of the final answer should be (e.g.\ the answer should be expressed in $cm^3$).\\
\indent \textbf{(Reformulation)} Next we reformulate the multiple choice question into an open-ended question by conditioning on the key extracted information extracted.\\
\indent \textbf{(Judgement)} Following the reformulation, a critical evaluation along multiple dimensions is conducted to evaluate whether the reformulation succeeded. This includes determining whether the new question is multiple choice, or if it even has a limited number of possible answers which would still be considered multiple choice. Additionally, we evaluate whether the reformulated problem is free from any information distortion: no existing information was changed, no critical information was omitted, and no new information was added. Finally, we determine whether the expected answer format is clear and if the problem can stand alone, without reference to the original options.\\
\indent \textbf{(Verification)} Finally, we perform a last verification that the entire process succeeded by checking that the expected information is present (e.g.\ expected answer format, rephrasing strategy, etc.).

\subsubsection{Reformulation Post-Processing}
The outcome of our reformulation is 88,983 questions that have passed the judgement and verification steps. However, we still need to ensure that the questions in \bigmath~satisfy the same criteria: uniquely verifiable, open-ended problems with closed-form solutions.
Therefore, we next examine the solvability of each reformulated problem by evaluating Llama-3.1-8B (8 rollouts) and Llama-3.1-405B (3 rollouts) on them. We filter the 88,983 problems down to only 48,698 by keeping only problems that were solved at least once by either model, but not solved 100\% by Llama-3.1-8B (to remove questions that may be too simple or obvious). In instances where neither model produces a solution that matches the reformulated answer, it is difficult to determine whether the issue lies with the model's performance, or with the reformulated answer. To mitigate such uncertainty, we exclude these problems from the dataset.

Finally, \bigmathmc~undergoes the same comprehensive filters as the rest of the datasets, as outlined in Section~\ref{sec:source_agnostic_filters}, yielding a final set of 47,010 reformulated and filtered problems, with the composition of source data found in Table~\ref{tab:bigmathmc}.
We find that this process has successfully reintroduced high-quality questions that were previously removed. Specifically, 72.7\% of the amc\_aime subset was found to be multiple choice, but the reformulation process successfully reintroduces 63.4\% of the multiple choice problems.


\section{Analysis and Discussion}
In this section we discuss and analyze the \bigmath~dataset. We consider the dataset difficulty, diversity, and outcomes of the filters that we proposed. Throughout the analysis, we include discussion points to aid the downstream uses of \bigmath.

\begin{figure}[t]
    \centering
    \includegraphics[width=0.95\linewidth]{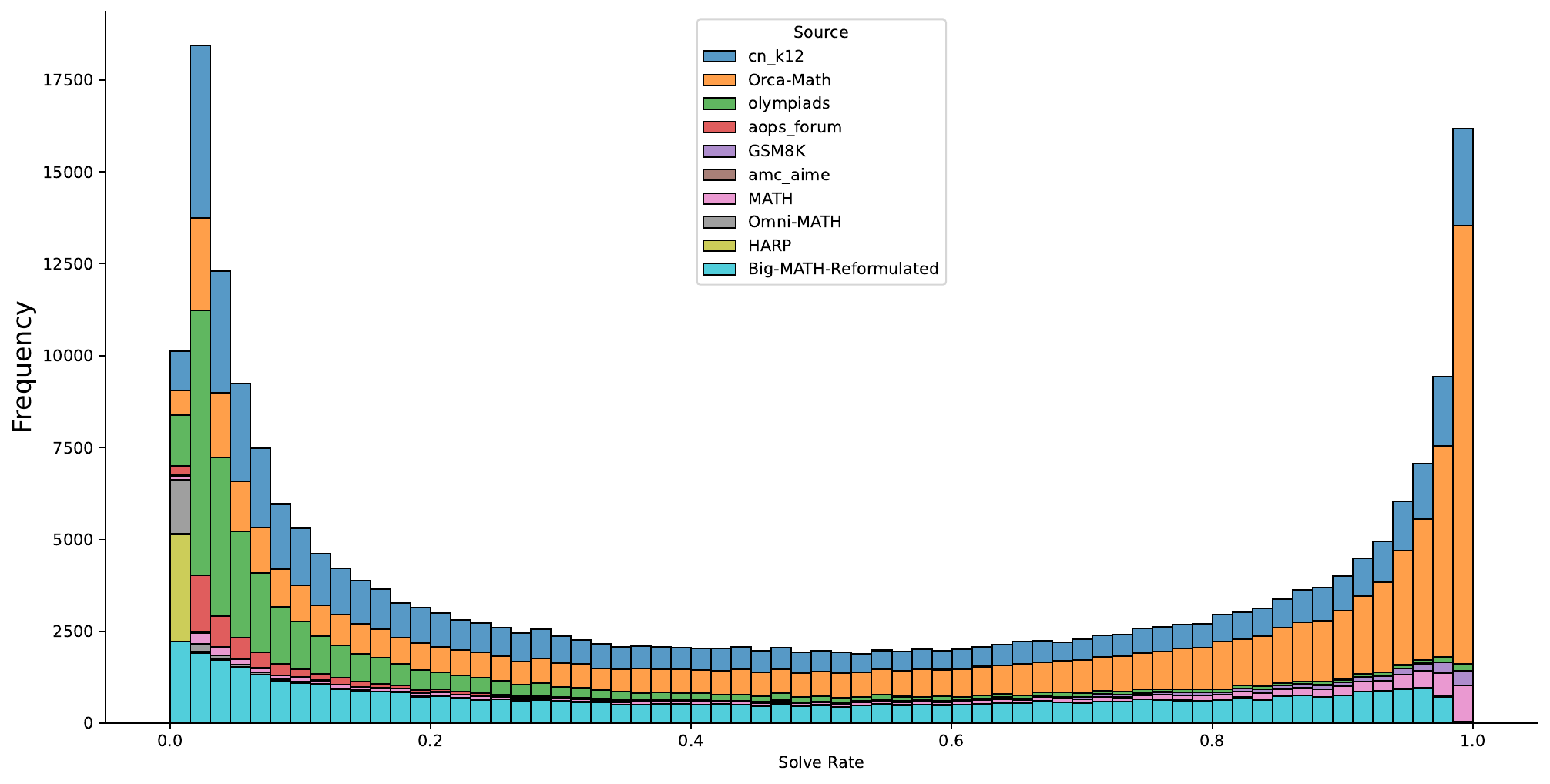}
    \caption{\textbf{Distribution of solve rates on each subset of \bigmath}, calculated with Llama-3.1-8B.}
    \label{fig:solve_rate_by_source}
\end{figure}

\begin{figure}[t]
    \centering
    \includegraphics[width=\linewidth]{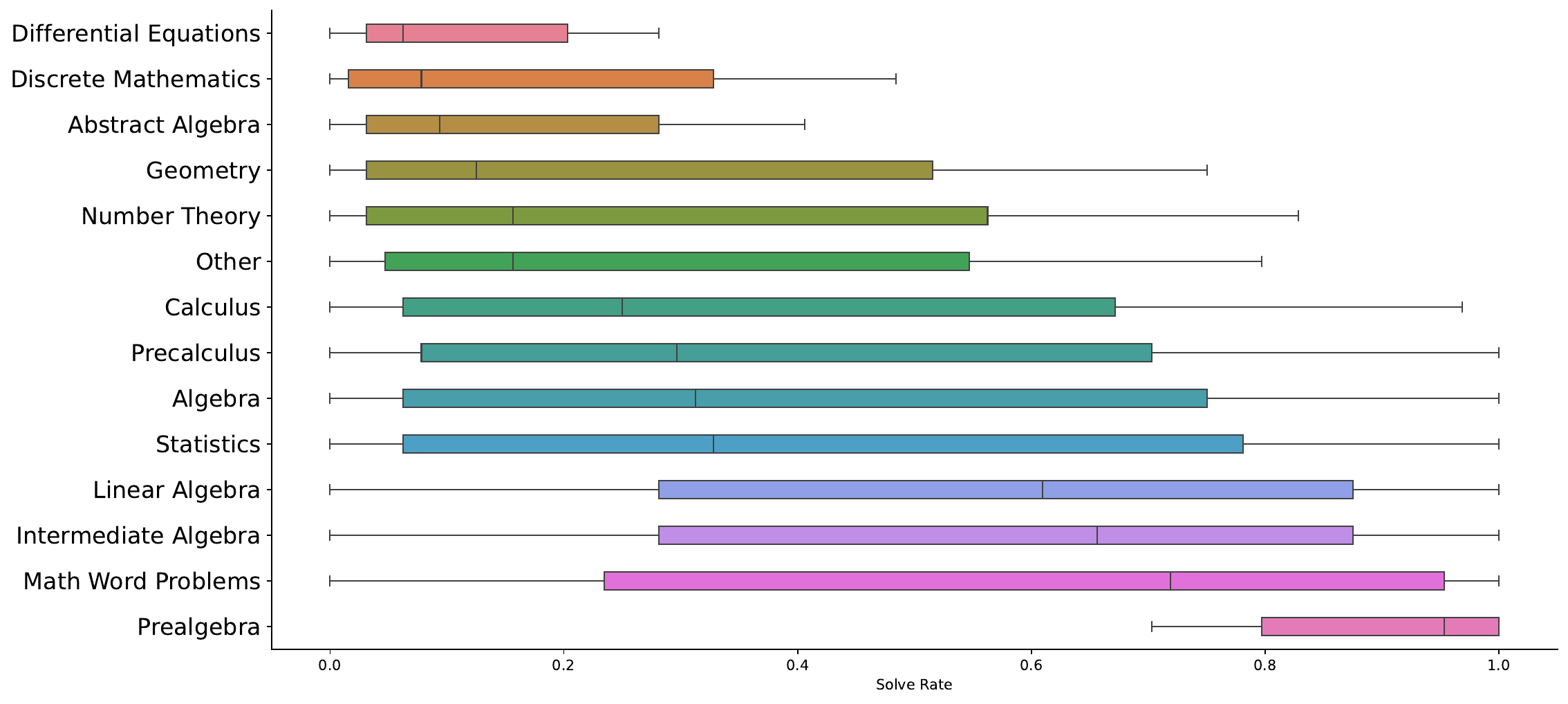}
    \caption{\textbf{Distribution of solve rates by domain}, calculated with Llama-3.1-8B.}
    \label{fig:solve_rate_by_domain}
\end{figure}

\subsection{Dataset Difficulty}
\label{sec:dataset_difficulty}
We calculate the difficulty of problems in \bigmath~based on rollouts from the Llama-3.1-8B model~\citep{dubey2024llama3herdmodels}, which provide a benchmark for understanding problem complexity.
For each problem in the dataset we generate 64 rollouts and calculate the success rate per problem. The distribution of success rates, split by source and domain is found in Figures~\ref{fig:solve_rate_by_source} and~\ref{fig:solve_rate_by_domain}.

First, Figure~\ref{fig:solve_rate_by_source} shows that the majority of the easiest data (highest solve rate) comes from the Orca-Math, cnk\_12, and MATH datasets, while the most difficult data is divided more evenly across the datasets. In particular, we find that nearly all of Omni-MATH and HARP are unsolveable by Llama-3.1-8B. Thus, in order to apply RL for Llama-3.1-8B on these difficult subsets, this particular model would need to be either be supervised fine-tuned on these datasets, or using an RL training algorithm that makes use of a process reward model. For example, Reinforcement Learning with Verifiable Rewards (RLVR)~\citep{lambert2024tulu3pushingfrontiers} would be unlikely to work effectively on Omni-Math and HARP as the models responses would produce no training signal.


Next, we group problems into difficulty quintiles, with the hardest quintile being problems that have a success rate less than 20\% and the easiest quintile with a success rate over 80\%. We find that, from easiest to hardest, the quintiles have 71,926 (28.64\%), 30,533 (12.16\%), 25,763 (10.26\%), 31,249 (12.44\%), and 91,647 problems (36.50\% of the total problems).
An obvious question now is: \textbf{how should practitioners use these dataset difficulties for their own purposes?}
In general, those training less capable, or smaller, models may want to remove the most difficult problems as it is unlikely that model rollouts will lead to a correct answer. This leads to inefficiency in the learning process because most RL methods used for LLMs (except those with a process reward model) will have 0 signal if the model can never come to the correct answer.
On the other hand, for those training a larger, or math-specific, model will find many of the easy questions redundant, and training on such data will be inefficient. Therefore, for practitioners training strong models it would be sensible to keep only the harder problems. Supposing that the hardest two quintiles of data are retained, there is still $> 120,000$ problems, \textit{10 times more problems than the next closest RL-suitable dataset.}

Next, we look at the difficulty of our novel \bigmathmc~subset. We see that our subset follows a similar solve rate distribution as the rest of the dataset; it has slightly more density around the low- and high-ends of the difficulty distribution. However, \bigmathmc~is skewed towards more difficult problems. Specifically, we find that 34.44\% of \bigmathmc~is in the hardest quintile, with an additional 16.42\% in the second hardest quintile, combining to greater than 50\% of the new data.

Finally, we look into the distribution of solve rates by each problem domain, shown in Figure~\ref{fig:solve_rate_by_domain}. We find that the most difficult problems come from the differential equations, discrete mathematics, and abstract algebra domains, while the prealgebra domain is the easiest by a wide margin. Interestingly, the remaining domains have a very wide distribution of difficulties, suggesting that within each domain there are likely problems requiring varying levels of expertise. Surprisingly, linear algebra was found to be one of the easier domains, while geometry was one of the most difficult domains, however, this may either be an artifact of the domain classification process, or of the specific training data for Llama-3.1-8B.

\begin{figure}
    \centering
    \includegraphics[width=0.95\linewidth]{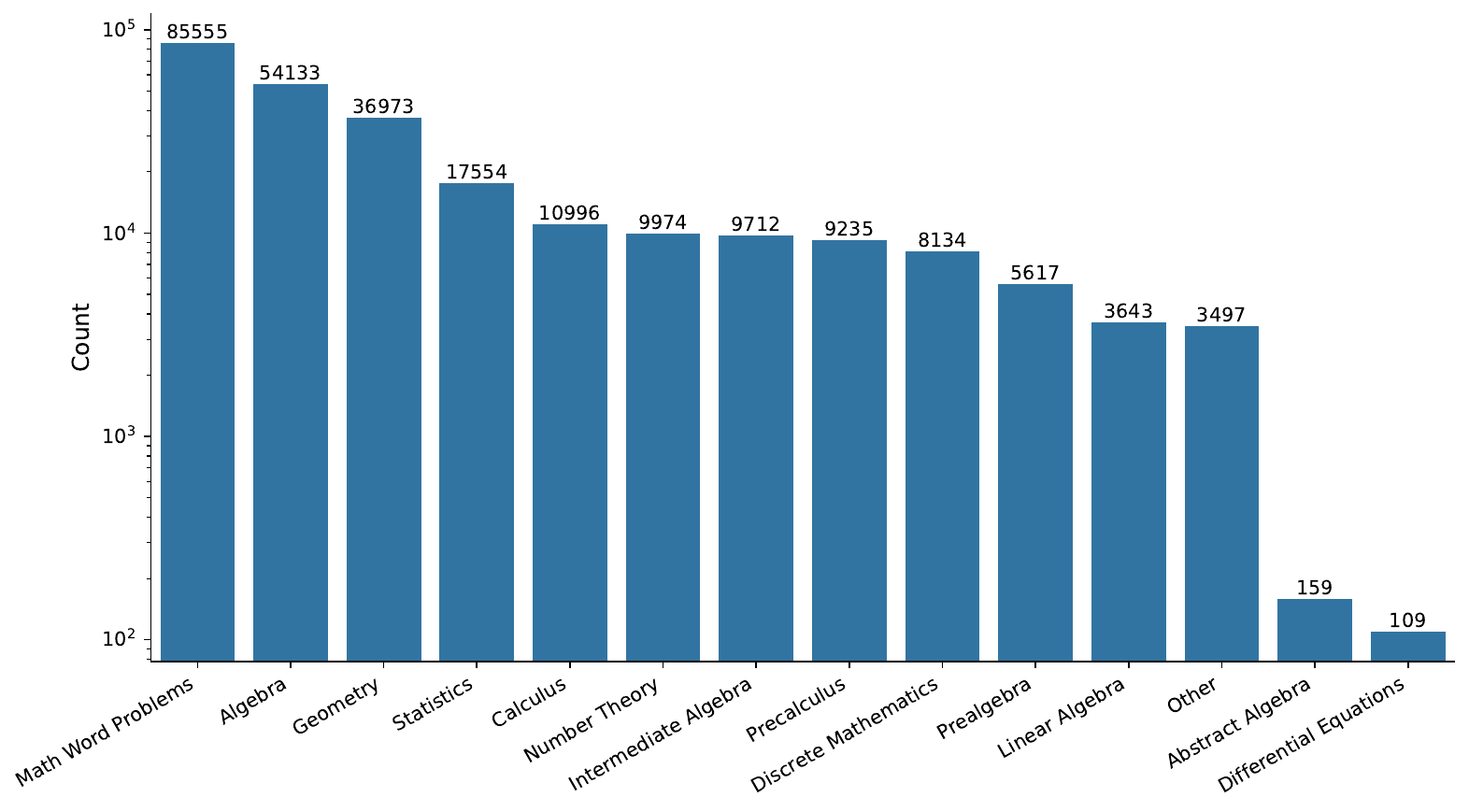}
    \caption{\textbf{Number of problems per domain}, as classified by gpt-4o-mini using math domains as defined by~\citet{gao2024omnimathuniversalolympiadlevel}. Note that the y-axis is log-scale.}
    \label{fig:omnimath_domains}
\end{figure}

\begin{figure}
    \centering
    \begin{text_full}{Example Problem: Math Word Problem}
    Question: A shopper buys a $100$ dollar coat on sale for $20\%$ off. An additional $5$ dollars are taken off the sale price by using a discount coupon. A sales tax of $8\%$ is paid on the final selling price. Calculate the total amount the shopper pays for the coat. Express your answer in dollars, rounded to two decimal places.\\
    Answer: 81.00
    \end{text_full}
    \begin{text_full}{Example Problem: Algebra}
    Question: Real numbers $x$ and $y$ satisfy $4x^{2}-5xy+4y^{2}=5$. If $s = x^{2} + y^{2}$, then $\frac{1}{s_{\max}} + \frac{1}{s_{\min}} = $?\\
    Answer: $\frac{8}{5}$
    \end{text_full}
    \begin{text_full}{Example Problem: Geometry}
    Question: Let  $ABCD$  be a cyclic quadrilateral, and suppose that  $BC = CD = 2$ . Let  $I$  be the incenter of triangle  $ABD$ . If  $AI = 2$  as well, find the minimum value of the length of diagonal  $BD$ .\\
    Answer: $2\sqrt{3}$
    \end{text_full}
    \begin{text_full}{Example Problem: Statistics}
    Question: A three-digit natural number $abc$ is termed a "convex number" if and only if the digits $a$, $b$, and $c$ (representing the hundreds, tens, and units place, respectively) satisfy $a < b$ and $c < b$. Given that $a$, $b$, and $c$ belong to the set $\{5, 6, 7, 8, 9\}$ and are distinct from one another, find the probability that a randomly chosen three-digit number $abc$ is a "convex number".\\
    Answer: $\frac{1}{3}$
    \end{text_full}
    \begin{text_full}{Example Problem: Calculus}
    Question: Find the sum $\frac{1}{2!}+\frac{2}{3!}+\frac{3}{4!}+\ldots+\frac{n}{(n+1)!}$ and compute its limit as $n \rightarrow \infty$.\\
    Answer: 1
    \end{text_full}
    \caption{\textbf{Example problems from the top-5 most common domains defined by~\citet{gao2024omnimathuniversalolympiadlevel}.}}
    \label{fig:omnimath_example_problems}
\end{figure}

\begin{figure}
    \centering
    \includegraphics[width=0.94\linewidth]{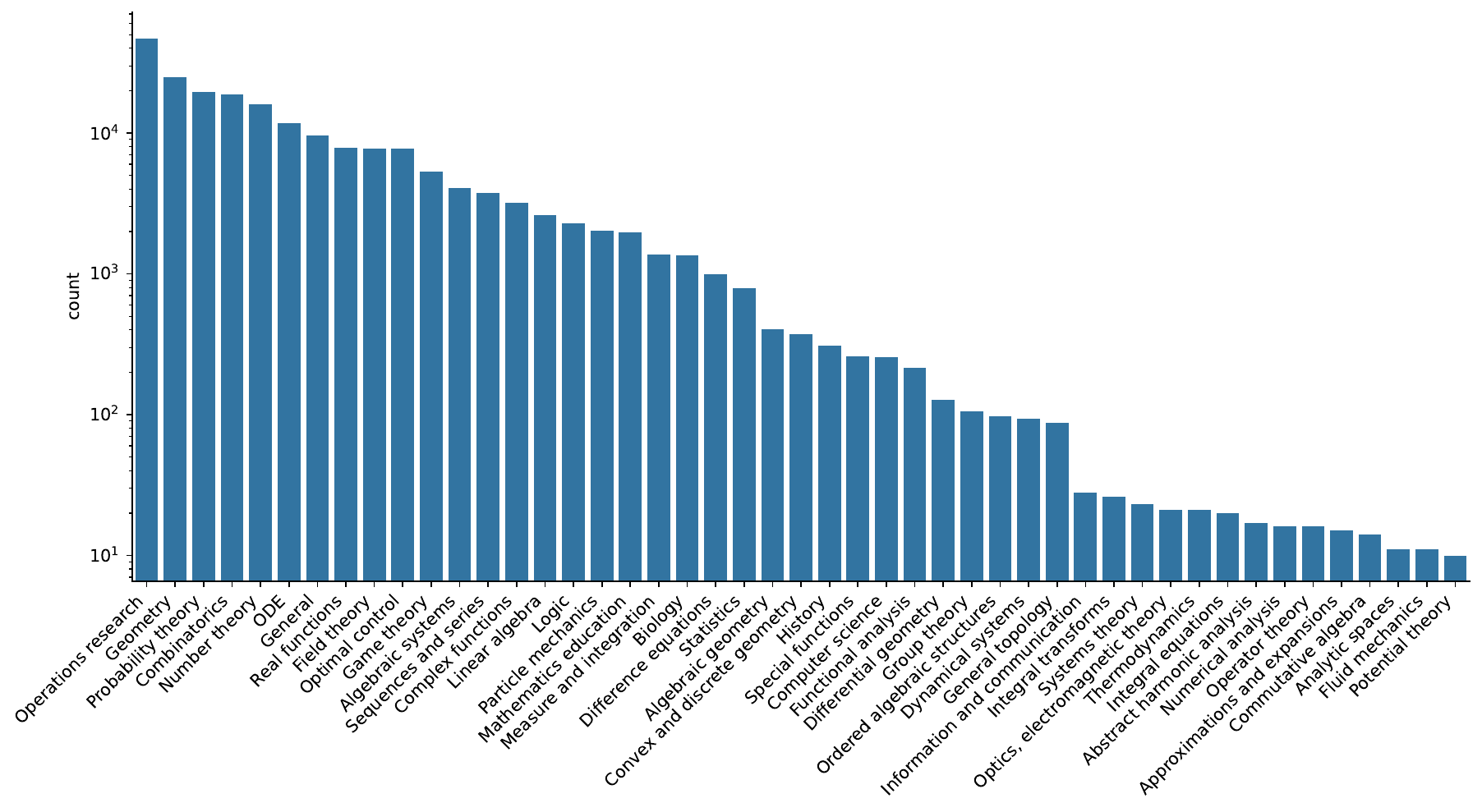}
    \caption{\textbf{Number of problems per domain}, as classified by gpt-4o-mini using math domains as defined by the 2020 Mathematics Subject Classification. Note that the y-axis is log-scale.}
    \label{fig:msc_domains}
\end{figure}

\begin{figure}
    \centering
    \begin{text_full}{Example Problem: Operations Research}
    Question: A book was sold at a profit of 12\%. Had it been sold for \$18 more, a profit of 18\% would have been gained. What is the cost price of the book?\\
    Answer: \$300
    \end{text_full}
    \begin{text_full}{Example Problem: Geometry}
    Question: Find the angles at which the parabola \( y = x^2 - x \) intersects the x-axis.\\
    Answer: $\frac{3\pi}{4} \text{ and } \frac{\pi}{4}$
    \end{text_full}
    \begin{text_full}{Example Problem: Probability Theory}
    Question: Ash and Gary independently come up with their own lineups of 15 fire, grass, and water monsters. Then, the first monster of both lineups will fight, with fire beating grass, grass beating water, and water beating fire. The defeated monster is then substituted with the next one from their team's lineup; if there is a draw, both monsters get defeated. Gary completes his lineup randomly, with each monster being equally likely to be any of the three types. Without seeing Gary's lineup, Ash chooses a lineup that maximizes the probability $p$ that his monsters are the last ones standing. Compute $p$.\\
    Answer: $1-\frac{2^{15}}{3^{15}}$
    \end{text_full}
    \begin{text_full}{Example Problem: Combinatorics}
    Question: Let $x_1\leq x_2\leq \cdots\leq x_{100}$ be real numbers such that $|x_1| + |x_2| + \cdots + |x_{100}| = 1$ and $x_1 + x_2 + \cdots + x_{100} = 0$. Among all such $100$-tuples of numbers, the greatest value that $x_{76} - x_{16}$ can achieve is $\tfrac mn$, where $m$ and $n$ are relatively prime positive integers. Find $m+n$.\\
    Answer: $841$
    \end{text_full}
    \begin{text_full}{Example Problem: Number Theory}
    Question: Suppose that $(u_n)$ is a sequence of real numbers satisfying
$u_{n+2}=2u_{n+1}+u_n$ and that $u_3=9$ and $u_6=128$. What is $u_5$?\\
    Answer: 53
    \end{text_full}
    \caption{\textbf{Example problems from the top-5 most common domains defined by the 2020 Mathematics Subject Classification.}}
    \label{fig:msc_example_problems}
\end{figure}

\subsection{Dataset Diversity}
\label{sec:dataset_diversity}
To better understand our dataset and the diversity of problems it provides, we next study the domain composition and distribution of our dataset.

We consider the distribution of problems sorted into mathematics domains according to two ontologies. The first ontology we consider is that proposed by~\citet{gao2024omnimathuniversalolympiadlevel}, and the second is the ontology defined by the American Mathematical Society in their 2020 Mathematics Subject Classification (MSC\footnote{\url{https://mathscinet.ams.org/mathscinet/msc/msc2020.html}}).
We use the same prompt and procedure as~\citet{gao2024omnimathuniversalolympiadlevel} to classify questions into their domains. 
Similar to~\citet{muennighoff2025s1simpletesttimescaling}, we classify problems into one of the 63 domains defined by the MSC using GPT-4o-mini.
The domains of~\citet{gao2024omnimathuniversalolympiadlevel} contain high-level topics ranging from grade-school to college-level, while the MSC domains are more fine-grained and include more detailed domains on math-adjacent topics.
For the domains defined by~\citet{gao2024omnimathuniversalolympiadlevel}, we present examples in Figure~\ref{fig:omnimath_example_problems} and a chart with the number of problems per domain in Figure~\ref{fig:omnimath_domains}, as well as a detailed breakdown of domains by source in Section~\ref{tab:domain_by_source} in the Appendix.
For the MSC domains, Figure~\ref{fig:omnimath_domains} shows the distribution of all domains with 10 or more problems, and Figure~\ref{fig:msc_example_problems} shows example problems for the top-5 most common domains.

We see in Figures~\ref{fig:omnimath_domains} and~\ref{fig:msc_domains} that the distributions for both ontologies have long tails. The largest domain from~\citet{gao2024omnimathuniversalolympiadlevel} is the Math Word Problems, which we find to come disproportionately from Orca-Math, which contains $>66,000$ such problems.
Surprisingly, we find that the largest MSC domain was operations research, and we inspected the data under this category to better understand where such a large quantity came from. We found that these problems are generally an application of algebra, geometry, or statistics to real world domains and could just as easily have been classified into another category.
We found that the distribution of MSC domains follows a nearly log-linear relation, with large quantities of problems in college-level topics, including $\sim10,000$ problems on ordinary differential equations, field theory, and optimal control. We also found smaller quantities of applications in the sciences in the dataset, with problems in electromagnetic theory, thermodynamics, and fluid mechanics, showing that the collected dataset contains problems from a wide variety of mathematics domains.

\subsection{Filter Analysis}
\label{sec:filter_analysis}
In this section we provide an analysis of the filtering process we designed to curate \bigmath.
See Table~\ref{tab:filter_result} for the main results and Appendix~\ref{app:filter_result_expanded} for detailed results.
Before looking closely at the filter outputs, it is important to note that our filtering process is intentionally designed to be more strict than is necessary. This conservative approach ensures that \bigmath~contains only the types of data that we desire. However, this leaves room for improvements to our filtering process, and expansion to the \bigmath~dataset. The filtering set-up proposed in this work serves only as a first step, allowing us to identify areas for improvement to maximize the utility of the source data and develop more precise filters.

In order to best identify where to spend future efforts, we start by identifying the filters which remove the most data: multiple-choice, multi-part, and proof filters.

\subsubsection{Multiple Choice Filter}
The largest portion of removed data falls under the multiple choice filter ($\sim 18\%$), with over 80\% of the multiple choice questions being sourced from the cn\_k12 subset. Since this filter affects such a large number of problems, we see this as promising area for improvement in the future, where the recall of filters can be improved to reintroduce some of the open-ended questions that are mistakenly classified as multiple choice. 
Additionally, we find large discrepancies in the classifications of the regular expression filter compared with the model-based filter, especially for the olympiad and MATH subsets, where the regular expression-based filter removes significantly more data than the model-based filter. A closer look at these data, which are classified as positive by one filter but negative by the other filter, are a prime area for further inspection. Exploration in this area can further improve the recall of the whole filtering system, and may lead to 1,000s of reintroduced problems.

\subsubsection{Multi-Part Filter}
Similarly, the multi-part filter shows potential for further refinement. Again, we see that the regular expression-based filter consistently removes a larger portion of data than the model-based filter. Specifically, we find that the regular expression-based filter removes 14,000 more questions from Orca-Math, 6,500 more from Olympiads, and 2,700 more from the MATH subset, compared with the model-based filter. Given these large differences, there is a clear opportunity to further improve this filtering step.

\subsubsection{Proof Filter}
Interestingly, the proof filter has the opposite outcome from the multiple choice and multi-part filters: the model-based filter removes 10,000 more problems than the regular expression-based filter. This is again an area for further investigation to determine how many of those problems can be reintroduced to the dataset, while still fulfilling our desiderata.

\subsubsection{Asymptote Filter}
One area that would be interesting to study further is a model's ability to use the Asymptote vector graphics language. In our work, we found 625 problems from the HARP subset using this language, and removed them all. However, this is not an insignificant portion of human-written problems, and may be beneficial to include these problems. In contrast to our work,~\citet{muennighoff2025s1simpletesttimescaling} evaluate a model on AIME24~\citep{aime} which contains images in the asymptote language. Presumably, training on such problems would be beneficial for this setting, however, they do not train a model on any asymptote images and do not analyze the performance of their model on such questions. Therefore, whether including such problems in the training set is beneficial is still unclear.



\section{Future Directions}
While this work primarily focuses on the creation and curation of the \bigmath and \bigmathmc datasets, this only represents the starting point for further research. Below we outline several promising directions for exploration.

\noindent \textbf{Using this dataset:}
\begin{itemize}
    \item \textbf{Scaling} - Investigate data scaling laws on RL training. In particular, there are a number of training algorithms, and new methods continually developing, and the efficiency of each is unknown~\citep{kimiteam2025kimik15scalingreinforcement}. For example, in very recent studies,~\citet{hou2025advancinglanguagemodelreasoning} demonstrate that training data and inference time compute scale in a complementary way (increasing inference is most useful when also scaling training data), and~\citet{setlur2025scalingtesttimecomputeverification} demonstrate that the gap between methods trained with verifier-based methods and verifier-free methods increases as the quantity of training data increases.
    \item \textbf{Balancing} - Future studies can analyze how to define and balance the quality, diversity, and complexity of a dataset~\citep{havrilla2024surveyingeffectsqualitydiversity}.
    \item \textbf{Leveraging difficulty} - Our difficulty classifications can be used studies on subset selection for RL, similar to the method proposed by~\citet{muennighoff2025s1simpletesttimescaling}. In particular, it will be interesting to explore how these difficulty scores translate to other models or model families. Additionally, a potentially fruitful avenue of research is to explore if difficulty scores can be used to form an effective curriculum, where curriculum learning has been effectively applied to RL in the past~\citep[inter alia]{dennis2021emergentcomplexityzeroshottransfer, jiang2021prioritizedlevelreplay}, but has only successfully been applied to training language models recently~\citep{albalak2023efficientonlinedatamixing, NEURIPS2024_ed165f2f}
    \item \textbf{Distillation} - As has been done with other datasets~\citep[inter alia]{liu2024acemathadvancingfrontiermath, guan2025rstarmathsmallllmsmaster}, \bigmath~an also be used in concert with a strong reasoning model to generate a dataset for SFT distillation.
\end{itemize}

\noindent \textbf{Improving future datasets:}
\begin{itemize}
    \item \textbf{Filter and verifier improvements} - As described throughout the paper, our filters are overly strict. This allows for further improvements on our filters, enabling the inclusion of more complex problems. Additionally, throughout this paper, we have assumed the use of a very simple verifier and only retained data that can be extracted based on simple string matching. However, the use of improved answer extraction and verifiers would allow for a much broader coverage of data, including proofs and multi-part questions.
    \item \textbf{Model-based filtering} - Refine and improve the process of designing model-based filters (e.g.\ prompting approaches). Integrating methods such as LLM-as-a-judge, or similar quality checks, to the model-based filtering can reduce the reliance on human oversight. Overall, the goal would be to develop advanced filters with fewer human-in-the-loop requirements. For instance, can RL training be used to train a filtering model, which in turn improves the RL training for a reasoning model? Probably! There exists huge quantities of positive and negative examples for the filters that we use here. The question then becomes: can the RL training teach a model to filter for new requirements? For example, training the model to identify data where the answer is not consistent with the model's solution. With methods such as Meta-RL or continual learning, this is a very exciting direction.
    \item \textbf{Domain expansion} - Some of the methodology in this work is specific to mathematics (e.g.\ proof filter), but much of the work is applicable to other domains (e.g.\ multiple choice and multi-part question filters). Extending the methodologies developed here to other domains is a very meaningful direction of study. A very recent example of generating verifiable question-answer pairs is that from~\citet{yuan2025naturalreasoningreasoningwild28m}, which can be used as inspiration.
\end{itemize}

\section{Conclusion}
There has been an incredible rise in the number of proposed methodologies on RL to train reasoning models~\citep{openai2024openaio1card, deepseek2024r1, liu2024acemathadvancingfrontiermath, yuan2024freeprocessrewardsprocess, guan2025rstarmathsmallllmsmaster}, where each method requires a large-scale corpus of high-quality, diverse prompts with verifiable answers.
While these recent works have proposed differing methodologies, many of them achieve similar results, suggesting that algorithms may not be the limiting factor in performance but, as has been proposed, data is the limiting factor on performance~\citep{xiang20252reasoningllmslearning, yeo2025demystifyinglongchainofthoughtreasoning}.

This work aims to address these issues by developing a large-scale corpus of high-quality, diverse math problems, with verifiable answers, \bigmath. Through our human-in-the-loop filtering process, we bridge the gap between data quality and quantity, leading to a dataset of over 250,000 problems satisfying our three key desiderata: uniquely verifiable solutions, open-ended formulations, and closed-form answers.
Furthermore, we introduce \bigmathmc, demonstrating the potential to reclaim valuable data lost during filtering, contributing over 47,000 high quality problems to the dataset. Through an in-depth analysis, we explored the diversity and difficulty of \bigmath, highlighting it's utility in advancing the state of the art in LLM reasoning capabilities, and pointing out promising directions for the future development of filters.

By releasing \bigmath~to the research community, we aim to catalyze progress in RL methodologies and establish a shared benchmark for future advancements in mathematical reasoning with LLMs. This dataset not only supports the development of more generalizable reasoning models but also lays the groundwork for exploring scaling laws, training dynamics, and new RL algorithms in the context of mathematical problem-solving.


\clearpage

\bibliography{main}

\begin{thebibliography}{48}
\providecommand{\natexlab}[1]{#1}
\providecommand{\url}[1]{\texttt{#1}}
\expandafter\ifx\csname urlstyle\endcsname\relax
  \providecommand{\doi}[1]{doi: #1}\else
  \providecommand{\doi}{doi: \begingroup \urlstyle{rm}\Url}\fi

\bibitem[Abbas et~al.(2023)Abbas, Tirumala, Simig, Ganguli, and Morcos]{abbas2023semdedupdataefficientlearningwebscale}
Amro Abbas, Kushal Tirumala, Dániel Simig, Surya Ganguli, and Ari~S. Morcos.
\newblock Semdedup: Data-efficient learning at web-scale through semantic deduplication, 2023.
\newblock URL \url{https://arxiv.org/abs/2303.09540}.

\bibitem[AI(2025)]{servicenow_r1distill}
ServiceNow AI.
\newblock Slam lab - r1-distill-sft dataset, 2025.
\newblock URL \url{https://hf.co/ServiceNow-AI/R1-Distill-SFT}.
\newblock Accessed: 2025-02-04.

\bibitem[Albalak et~al.(2023)Albalak, Pan, Raffel, and Wang]{albalak2023efficientonlinedatamixing}
Alon Albalak, Liangming Pan, Colin Raffel, and William~Yang Wang.
\newblock Efficient online data mixing for language model pre-training, 2023.
\newblock URL \url{https://arxiv.org/abs/2312.02406}.

\bibitem[Albalak et~al.(2024)Albalak, Elazar, Xie, Longpre, Lambert, Wang, Muennighoff, Hou, Pan, Jeong, Raffel, Chang, Hashimoto, and Wang]{albalak2024surveydataselectionlanguage}
Alon Albalak, Yanai Elazar, Sang~Michael Xie, Shayne Longpre, Nathan Lambert, Xinyi Wang, Niklas Muennighoff, Bairu Hou, Liangming Pan, Haewon Jeong, Colin Raffel, Shiyu Chang, Tatsunori Hashimoto, and William~Yang Wang.
\newblock A survey on data selection for language models, 2024.
\newblock URL \url{https://arxiv.org/abs/2402.16827}.

\bibitem[Chu et~al.(2025)Chu, Zhai, Yang, Tong, Xie, Schuurmans, Le, Levine, and Ma]{chu2025sftmemorizesrlgeneralizes}
Tianzhe Chu, Yuexiang Zhai, Jihan Yang, Shengbang Tong, Saining Xie, Dale Schuurmans, Quoc~V. Le, Sergey Levine, and Yi~Ma.
\newblock Sft memorizes, rl generalizes: A comparative study of foundation model post-training, 2025.
\newblock URL \url{https://arxiv.org/abs/2501.17161}.

\bibitem[Cobbe et~al.(2021)Cobbe, Kosaraju, Bavarian, Chen, Jun, Kaiser, Plappert, Tworek, Hilton, Nakano, Hesse, and Schulman]{cobbe2021trainingverifierssolvemath}
Karl Cobbe, Vineet Kosaraju, Mohammad Bavarian, Mark Chen, Heewoo Jun, Lukasz Kaiser, Matthias Plappert, Jerry Tworek, Jacob Hilton, Reiichiro Nakano, Christopher Hesse, and John Schulman.
\newblock Training verifiers to solve math word problems, 2021.
\newblock URL \url{https://arxiv.org/abs/2110.14168}.

\bibitem[DeepSeek-AI et~al.(2025)DeepSeek-AI, Guo, Yang, Zhang, Song, Zhang, Xu, Zhu, Ma, Wang, Bi, Zhang, Yu, Wu, Wu, Gou, Shao, Li, Gao, Liu, Xue, Wang, Wu, Feng, Lu, Zhao, Deng, Zhang, Ruan, Dai, Chen, Ji, Li, Lin, Dai, Luo, Hao, Chen, Li, Zhang, Bao, Xu, Wang, Ding, Xin, Gao, Qu, Li, Guo, Li, Wang, Chen, Yuan, Qiu, Li, Cai, Ni, Liang, Chen, Dong, Hu, Gao, Guan, Huang, Yu, Wang, Zhang, Zhao, Wang, Zhang, Xu, Xia, Zhang, Zhang, Tang, Li, Wang, Li, Tian, Huang, Zhang, Wang, Chen, Du, Ge, Zhang, Pan, Wang, Chen, Jin, Chen, Lu, Zhou, Chen, Ye, Wang, Yu, Zhou, Pan, Li, Zhou, Wu, Ye, Yun, Pei, Sun, Wang, Zeng, Zhao, Liu, Liang, Gao, Yu, Zhang, Xiao, An, Liu, Wang, Chen, Nie, Cheng, Liu, Xie, Liu, Yang, Li, Su, Lin, Li, Jin, Shen, Chen, Sun, Wang, Song, Zhou, Wang, Shan, Li, Wang, Wei, Zhang, Xu, Li, Zhao, Sun, Wang, Yu, Zhang, Shi, Xiong, He, Piao, Wang, Tan, Ma, Liu, Guo, Ou, Wang, Gong, Zou, He, Xiong, Luo, You, Liu, Zhou, Zhu, Xu, Huang, Li, Zheng, Zhu, Ma, Tang, Zha, Yan, Ren, Ren, Sha, Fu, Xu, Xie, Zhang,
  Hao, Ma, Yan, Wu, Gu, Zhu, Liu, Li, Xie, Song, Pan, Huang, Xu, Zhang, and Zhang]{deepseek2024r1}
DeepSeek-AI, Daya Guo, Dejian Yang, Haowei Zhang, Junxiao Song, Ruoyu Zhang, Runxin Xu, Qihao Zhu, Shirong Ma, Peiyi Wang, Xiao Bi, Xiaokang Zhang, Xingkai Yu, Yu~Wu, Z.~F. Wu, Zhibin Gou, Zhihong Shao, Zhuoshu Li, Ziyi Gao, Aixin Liu, Bing Xue, Bingxuan Wang, Bochao Wu, Bei Feng, Chengda Lu, Chenggang Zhao, Chengqi Deng, Chenyu Zhang, Chong Ruan, Damai Dai, Deli Chen, Dongjie Ji, Erhang Li, Fangyun Lin, Fucong Dai, Fuli Luo, Guangbo Hao, Guanting Chen, Guowei Li, H.~Zhang, Han Bao, Hanwei Xu, Haocheng Wang, Honghui Ding, Huajian Xin, Huazuo Gao, Hui Qu, Hui Li, Jianzhong Guo, Jiashi Li, Jiawei Wang, Jingchang Chen, Jingyang Yuan, Junjie Qiu, Junlong Li, J.~L. Cai, Jiaqi Ni, Jian Liang, Jin Chen, Kai Dong, Kai Hu, Kaige Gao, Kang Guan, Kexin Huang, Kuai Yu, Lean Wang, Lecong Zhang, Liang Zhao, Litong Wang, Liyue Zhang, Lei Xu, Leyi Xia, Mingchuan Zhang, Minghua Zhang, Minghui Tang, Meng Li, Miaojun Wang, Mingming Li, Ning Tian, Panpan Huang, Peng Zhang, Qiancheng Wang, Qinyu Chen, Qiushi Du, Ruiqi Ge, Ruisong
  Zhang, Ruizhe Pan, Runji Wang, R.~J. Chen, R.~L. Jin, Ruyi Chen, Shanghao Lu, Shangyan Zhou, Shanhuang Chen, Shengfeng Ye, Shiyu Wang, Shuiping Yu, Shunfeng Zhou, Shuting Pan, S.~S. Li, Shuang Zhou, Shaoqing Wu, Shengfeng Ye, Tao Yun, Tian Pei, Tianyu Sun, T.~Wang, Wangding Zeng, Wanjia Zhao, Wen Liu, Wenfeng Liang, Wenjun Gao, Wenqin Yu, Wentao Zhang, W.~L. Xiao, Wei An, Xiaodong Liu, Xiaohan Wang, Xiaokang Chen, Xiaotao Nie, Xin Cheng, Xin Liu, Xin Xie, Xingchao Liu, Xinyu Yang, Xinyuan Li, Xuecheng Su, Xuheng Lin, X.~Q. Li, Xiangyue Jin, Xiaojin Shen, Xiaosha Chen, Xiaowen Sun, Xiaoxiang Wang, Xinnan Song, Xinyi Zhou, Xianzu Wang, Xinxia Shan, Y.~K. Li, Y.~Q. Wang, Y.~X. Wei, Yang Zhang, Yanhong Xu, Yao Li, Yao Zhao, Yaofeng Sun, Yaohui Wang, Yi~Yu, Yichao Zhang, Yifan Shi, Yiliang Xiong, Ying He, Yishi Piao, Yisong Wang, Yixuan Tan, Yiyang Ma, Yiyuan Liu, Yongqiang Guo, Yuan Ou, Yuduan Wang, Yue Gong, Yuheng Zou, Yujia He, Yunfan Xiong, Yuxiang Luo, Yuxiang You, Yuxuan Liu, Yuyang Zhou, Y.~X. Zhu,
  Yanhong Xu, Yanping Huang, Yaohui Li, Yi~Zheng, Yuchen Zhu, Yunxian Ma, Ying Tang, Yukun Zha, Yuting Yan, Z.~Z. Ren, Zehui Ren, Zhangli Sha, Zhe Fu, Zhean Xu, Zhenda Xie, Zhengyan Zhang, Zhewen Hao, Zhicheng Ma, Zhigang Yan, Zhiyu Wu, Zihui Gu, Zijia Zhu, Zijun Liu, Zilin Li, Ziwei Xie, Ziyang Song, Zizheng Pan, Zhen Huang, Zhipeng Xu, Zhongyu Zhang, and Zhen Zhang.
\newblock Deepseek-r1: Incentivizing reasoning capability in llms via reinforcement learning, 2025.
\newblock URL \url{https://arxiv.org/abs/2501.12948}.

\bibitem[Dennis et~al.(2021)Dennis, Jaques, Vinitsky, Bayen, Russell, Critch, and Levine]{dennis2021emergentcomplexityzeroshottransfer}
Michael Dennis, Natasha Jaques, Eugene Vinitsky, Alexandre Bayen, Stuart Russell, Andrew Critch, and Sergey Levine.
\newblock Emergent complexity and zero-shot transfer via unsupervised environment design, 2021.
\newblock URL \url{https://arxiv.org/abs/2012.02096}.

\bibitem[Dubey et~al.(2024)Dubey, Jauhri, Pandey, Kadian, Al-Dahle, Letman, Mathur, Schelten, Yang, Fan, Goyal, Hartshorn, Yang, Mitra, Sravankumar, Korenev, Hinsvark, Rao, Zhang, Rodriguez, Gregerson, Spataru, Roziere, Biron, Tang, Chern, Caucheteux, Nayak, Bi, Marra, McConnell, Keller, Touret, Wu, Wong, Ferrer, Nikolaidis, Allonsius, Song, Pintz, Livshits, Esiobu, Choudhary, Mahajan, Garcia-Olano, Perino, Hupkes, Lakomkin, AlBadawy, Lobanova, Dinan, Smith, Radenovic, Zhang, Synnaeve, Lee, Anderson, Nail, Mialon, Pang, Cucurell, Nguyen, Korevaar, Xu, Touvron, Zarov, Ibarra, Kloumann, Misra, Evtimov, Copet, Lee, Geffert, Vranes, Park, Mahadeokar, Shah, van~der Linde, Billock, Hong, Lee, Fu, Chi, Huang, Liu, Wang, Yu, Bitton, Spisak, Park, Rocca, Johnstun, Saxe, Jia, Alwala, Upasani, Plawiak, Li, Heafield, Stone, El-Arini, Iyer, Malik, Chiu, Bhalla, Rantala-Yeary, van~der Maaten, Chen, Tan, Jenkins, Martin, Madaan, Malo, Blecher, Landzaat, de~Oliveira, Muzzi, Pasupuleti, Singh, Paluri, Kardas, Oldham, Rita,
  Pavlova, Kambadur, Lewis, Si, Singh, Hassan, Goyal, Torabi, Bashlykov, Bogoychev, Chatterji, Duchenne, Çelebi, Alrassy, Zhang, Li, Vasic, Weng, Bhargava, Dubal, Krishnan, Koura, Xu, He, Dong, Srinivasan, Ganapathy, Calderer, Cabral, Stojnic, Raileanu, Girdhar, Patel, Sauvestre, Polidoro, Sumbaly, Taylor, Silva, Hou, Wang, Hosseini, Chennabasappa, Singh, Bell, Kim, Edunov, Nie, Narang, Raparthy, Shen, Wan, Bhosale, Zhang, Vandenhende, Batra, Whitman, Sootla, Collot, Gururangan, Borodinsky, Herman, Fowler, Sheasha, Georgiou, Scialom, Speckbacher, Mihaylov, Xiao, Karn, Goswami, Gupta, Ramanathan, Kerkez, Gonguet, Do, Vogeti, Petrovic, Chu, Xiong, Fu, Meers, Martinet, Wang, Tan, Xie, Jia, Wang, Goldschlag, Gaur, Babaei, Wen, Song, Zhang, Li, Mao, Coudert, Yan, Chen, Papakipos, Singh, Grattafiori, Jain, Kelsey, Shajnfeld, Gangidi, Victoria, Goldstand, Menon, Sharma, Boesenberg, Vaughan, Baevski, Feinstein, Kallet, Sangani, Yunus, Lupu, Alvarado, Caples, Gu, Ho, Poulton, Ryan, Ramchandani, Franco, Saraf,
  Chowdhury, Gabriel, Bharambe, Eisenman, Yazdan, James, Maurer, Leonhardi, Huang, Loyd, Paola, Paranjape, Liu, Wu, Ni, Hancock, Wasti, Spence, Stojkovic, Gamido, Montalvo, Parker, Burton, Mejia, Wang, Kim, Zhou, Hu, Chu, Cai, Tindal, Feichtenhofer, Civin, Beaty, Kreymer, Li, Wyatt, Adkins, Xu, Testuggine, David, Parikh, Liskovich, Foss, Wang, Le, Holland, Dowling, Jamil, Montgomery, Presani, Hahn, Wood, Brinkman, Arcaute, Dunbar, Smothers, Sun, Kreuk, Tian, Ozgenel, Caggioni, Guzmán, Kanayet, Seide, Florez, Schwarz, Badeer, Swee, Halpern, Thattai, Herman, Sizov, Guangyi, Zhang, Lakshminarayanan, Shojanazeri, Zou, Wang, Zha, Habeeb, Rudolph, Suk, Aspegren, Goldman, Damlaj, Molybog, Tufanov, Veliche, Gat, Weissman, Geboski, Kohli, Asher, Gaya, Marcus, Tang, Chan, Zhen, Reizenstein, Teboul, Zhong, Jin, Yang, Cummings, Carvill, Shepard, McPhie, Torres, Ginsburg, Wang, Wu, U, Saxena, Prasad, Khandelwal, Zand, Matosich, Veeraraghavan, Michelena, Li, Huang, Chawla, Lakhotia, Huang, Chen, Garg, A, Silva, Bell,
  Zhang, Guo, Yu, Moshkovich, Wehrstedt, Khabsa, Avalani, Bhatt, Tsimpoukelli, Mankus, Hasson, Lennie, Reso, Groshev, Naumov, Lathi, Keneally, Seltzer, Valko, Restrepo, Patel, Vyatskov, Samvelyan, Clark, Macey, Wang, Hermoso, Metanat, Rastegari, Bansal, Santhanam, Parks, White, Bawa, Singhal, Egebo, Usunier, Laptev, Dong, Zhang, Cheng, Chernoguz, Hart, Salpekar, Kalinli, Kent, Parekh, Saab, Balaji, Rittner, Bontrager, Roux, Dollar, Zvyagina, Ratanchandani, Yuvraj, Liang, Alao, Rodriguez, Ayub, Murthy, Nayani, Mitra, Li, Hogan, Battey, Wang, Maheswari, Howes, Rinott, Bondu, Datta, Chugh, Hunt, Dhillon, Sidorov, Pan, Verma, Yamamoto, Ramaswamy, Lindsay, Lindsay, Feng, Lin, Zha, Shankar, Zhang, Zhang, Wang, Agarwal, Sajuyigbe, Chintala, Max, Chen, Kehoe, Satterfield, Govindaprasad, Gupta, Cho, Virk, Subramanian, Choudhury, Goldman, Remez, Glaser, Best, Kohler, Robinson, Li, Zhang, Matthews, Chou, Shaked, Vontimitta, Ajayi, Montanez, Mohan, Kumar, Mangla, Albiero, Ionescu, Poenaru, Mihailescu, Ivanov, Li, Wang,
  Jiang, Bouaziz, Constable, Tang, Wang, Wu, Wang, Xia, Wu, Gao, Chen, Hu, Jia, Qi, Li, Zhang, Zhang, Adi, Nam, Yu, Wang, Hao, Qian, He, Rait, DeVito, Rosnbrick, Wen, Yang, and Zhao]{dubey2024llama3herdmodels}
Abhimanyu Dubey, Abhinav Jauhri, Abhinav Pandey, Abhishek Kadian, Ahmad Al-Dahle, Aiesha Letman, Akhil Mathur, Alan Schelten, Amy Yang, Angela Fan, Anirudh Goyal, Anthony Hartshorn, Aobo Yang, Archi Mitra, Archie Sravankumar, Artem Korenev, Arthur Hinsvark, Arun Rao, Aston Zhang, Aurelien Rodriguez, Austen Gregerson, Ava Spataru, Baptiste Roziere, Bethany Biron, Binh Tang, Bobbie Chern, Charlotte Caucheteux, Chaya Nayak, Chloe Bi, Chris Marra, Chris McConnell, Christian Keller, Christophe Touret, Chunyang Wu, Corinne Wong, Cristian~Canton Ferrer, Cyrus Nikolaidis, Damien Allonsius, Daniel Song, Danielle Pintz, Danny Livshits, David Esiobu, Dhruv Choudhary, Dhruv Mahajan, Diego Garcia-Olano, Diego Perino, Dieuwke Hupkes, Egor Lakomkin, Ehab AlBadawy, Elina Lobanova, Emily Dinan, Eric~Michael Smith, Filip Radenovic, Frank Zhang, Gabriel Synnaeve, Gabrielle Lee, Georgia~Lewis Anderson, Graeme Nail, Gregoire Mialon, Guan Pang, Guillem Cucurell, Hailey Nguyen, Hannah Korevaar, Hu~Xu, Hugo Touvron, Iliyan Zarov,
  Imanol~Arrieta Ibarra, Isabel Kloumann, Ishan Misra, Ivan Evtimov, Jade Copet, Jaewon Lee, Jan Geffert, Jana Vranes, Jason Park, Jay Mahadeokar, Jeet Shah, Jelmer van~der Linde, Jennifer Billock, Jenny Hong, Jenya Lee, Jeremy Fu, Jianfeng Chi, Jianyu Huang, Jiawen Liu, Jie Wang, Jiecao Yu, Joanna Bitton, Joe Spisak, Jongsoo Park, Joseph Rocca, Joshua Johnstun, Joshua Saxe, Junteng Jia, Kalyan~Vasuden Alwala, Kartikeya Upasani, Kate Plawiak, Ke~Li, Kenneth Heafield, Kevin Stone, Khalid El-Arini, Krithika Iyer, Kshitiz Malik, Kuenley Chiu, Kunal Bhalla, Lauren Rantala-Yeary, Laurens van~der Maaten, Lawrence Chen, Liang Tan, Liz Jenkins, Louis Martin, Lovish Madaan, Lubo Malo, Lukas Blecher, Lukas Landzaat, Luke de~Oliveira, Madeline Muzzi, Mahesh Pasupuleti, Mannat Singh, Manohar Paluri, Marcin Kardas, Mathew Oldham, Mathieu Rita, Maya Pavlova, Melanie Kambadur, Mike Lewis, Min Si, Mitesh~Kumar Singh, Mona Hassan, Naman Goyal, Narjes Torabi, Nikolay Bashlykov, Nikolay Bogoychev, Niladri Chatterji, Olivier
  Duchenne, Onur Çelebi, Patrick Alrassy, Pengchuan Zhang, Pengwei Li, Petar Vasic, Peter Weng, Prajjwal Bhargava, Pratik Dubal, Praveen Krishnan, Punit~Singh Koura, Puxin Xu, Qing He, Qingxiao Dong, Ragavan Srinivasan, Raj Ganapathy, Ramon Calderer, Ricardo~Silveira Cabral, Robert Stojnic, Roberta Raileanu, Rohit Girdhar, Rohit Patel, Romain Sauvestre, Ronnie Polidoro, Roshan Sumbaly, Ross Taylor, Ruan Silva, Rui Hou, Rui Wang, Saghar Hosseini, Sahana Chennabasappa, Sanjay Singh, Sean Bell, Seohyun~Sonia Kim, Sergey Edunov, Shaoliang Nie, Sharan Narang, Sharath Raparthy, Sheng Shen, Shengye Wan, Shruti Bhosale, Shun Zhang, Simon Vandenhende, Soumya Batra, Spencer Whitman, Sten Sootla, Stephane Collot, Suchin Gururangan, Sydney Borodinsky, Tamar Herman, Tara Fowler, Tarek Sheasha, Thomas Georgiou, Thomas Scialom, Tobias Speckbacher, Todor Mihaylov, Tong Xiao, Ujjwal Karn, Vedanuj Goswami, Vibhor Gupta, Vignesh Ramanathan, Viktor Kerkez, Vincent Gonguet, Virginie Do, Vish Vogeti, Vladan Petrovic, Weiwei Chu,
  Wenhan Xiong, Wenyin Fu, Whitney Meers, Xavier Martinet, Xiaodong Wang, Xiaoqing~Ellen Tan, Xinfeng Xie, Xuchao Jia, Xuewei Wang, Yaelle Goldschlag, Yashesh Gaur, Yasmine Babaei, Yi~Wen, Yiwen Song, Yuchen Zhang, Yue Li, Yuning Mao, Zacharie~Delpierre Coudert, Zheng Yan, Zhengxing Chen, Zoe Papakipos, Aaditya Singh, Aaron Grattafiori, Abha Jain, Adam Kelsey, Adam Shajnfeld, Adithya Gangidi, Adolfo Victoria, Ahuva Goldstand, Ajay Menon, Ajay Sharma, Alex Boesenberg, Alex Vaughan, Alexei Baevski, Allie Feinstein, Amanda Kallet, Amit Sangani, Anam Yunus, Andrei Lupu, Andres Alvarado, Andrew Caples, Andrew Gu, Andrew Ho, Andrew Poulton, Andrew Ryan, Ankit Ramchandani, Annie Franco, Aparajita Saraf, Arkabandhu Chowdhury, Ashley Gabriel, Ashwin Bharambe, Assaf Eisenman, Azadeh Yazdan, Beau James, Ben Maurer, Benjamin Leonhardi, Bernie Huang, Beth Loyd, Beto~De Paola, Bhargavi Paranjape, Bing Liu, Bo~Wu, Boyu Ni, Braden Hancock, Bram Wasti, Brandon Spence, Brani Stojkovic, Brian Gamido, Britt Montalvo, Carl
  Parker, Carly Burton, Catalina Mejia, Changhan Wang, Changkyu Kim, Chao Zhou, Chester Hu, Ching-Hsiang Chu, Chris Cai, Chris Tindal, Christoph Feichtenhofer, Damon Civin, Dana Beaty, Daniel Kreymer, Daniel Li, Danny Wyatt, David Adkins, David Xu, Davide Testuggine, Delia David, Devi Parikh, Diana Liskovich, Didem Foss, Dingkang Wang, Duc Le, Dustin Holland, Edward Dowling, Eissa Jamil, Elaine Montgomery, Eleonora Presani, Emily Hahn, Emily Wood, Erik Brinkman, Esteban Arcaute, Evan Dunbar, Evan Smothers, Fei Sun, Felix Kreuk, Feng Tian, Firat Ozgenel, Francesco Caggioni, Francisco Guzmán, Frank Kanayet, Frank Seide, Gabriela~Medina Florez, Gabriella Schwarz, Gada Badeer, Georgia Swee, Gil Halpern, Govind Thattai, Grant Herman, Grigory Sizov, Guangyi, Zhang, Guna Lakshminarayanan, Hamid Shojanazeri, Han Zou, Hannah Wang, Hanwen Zha, Haroun Habeeb, Harrison Rudolph, Helen Suk, Henry Aspegren, Hunter Goldman, Ibrahim Damlaj, Igor Molybog, Igor Tufanov, Irina-Elena Veliche, Itai Gat, Jake Weissman, James
  Geboski, James Kohli, Japhet Asher, Jean-Baptiste Gaya, Jeff Marcus, Jeff Tang, Jennifer Chan, Jenny Zhen, Jeremy Reizenstein, Jeremy Teboul, Jessica Zhong, Jian Jin, Jingyi Yang, Joe Cummings, Jon Carvill, Jon Shepard, Jonathan McPhie, Jonathan Torres, Josh Ginsburg, Junjie Wang, Kai Wu, Kam~Hou U, Karan Saxena, Karthik Prasad, Kartikay Khandelwal, Katayoun Zand, Kathy Matosich, Kaushik Veeraraghavan, Kelly Michelena, Keqian Li, Kun Huang, Kunal Chawla, Kushal Lakhotia, Kyle Huang, Lailin Chen, Lakshya Garg, Lavender A, Leandro Silva, Lee Bell, Lei Zhang, Liangpeng Guo, Licheng Yu, Liron Moshkovich, Luca Wehrstedt, Madian Khabsa, Manav Avalani, Manish Bhatt, Maria Tsimpoukelli, Martynas Mankus, Matan Hasson, Matthew Lennie, Matthias Reso, Maxim Groshev, Maxim Naumov, Maya Lathi, Meghan Keneally, Michael~L. Seltzer, Michal Valko, Michelle Restrepo, Mihir Patel, Mik Vyatskov, Mikayel Samvelyan, Mike Clark, Mike Macey, Mike Wang, Miquel~Jubert Hermoso, Mo~Metanat, Mohammad Rastegari, Munish Bansal, Nandhini
  Santhanam, Natascha Parks, Natasha White, Navyata Bawa, Nayan Singhal, Nick Egebo, Nicolas Usunier, Nikolay~Pavlovich Laptev, Ning Dong, Ning Zhang, Norman Cheng, Oleg Chernoguz, Olivia Hart, Omkar Salpekar, Ozlem Kalinli, Parkin Kent, Parth Parekh, Paul Saab, Pavan Balaji, Pedro Rittner, Philip Bontrager, Pierre Roux, Piotr Dollar, Polina Zvyagina, Prashant Ratanchandani, Pritish Yuvraj, Qian Liang, Rachad Alao, Rachel Rodriguez, Rafi Ayub, Raghotham Murthy, Raghu Nayani, Rahul Mitra, Raymond Li, Rebekkah Hogan, Robin Battey, Rocky Wang, Rohan Maheswari, Russ Howes, Ruty Rinott, Sai~Jayesh Bondu, Samyak Datta, Sara Chugh, Sara Hunt, Sargun Dhillon, Sasha Sidorov, Satadru Pan, Saurabh Verma, Seiji Yamamoto, Sharadh Ramaswamy, Shaun Lindsay, Shaun Lindsay, Sheng Feng, Shenghao Lin, Shengxin~Cindy Zha, Shiva Shankar, Shuqiang Zhang, Shuqiang Zhang, Sinong Wang, Sneha Agarwal, Soji Sajuyigbe, Soumith Chintala, Stephanie Max, Stephen Chen, Steve Kehoe, Steve Satterfield, Sudarshan Govindaprasad, Sumit Gupta,
  Sungmin Cho, Sunny Virk, Suraj Subramanian, Sy~Choudhury, Sydney Goldman, Tal Remez, Tamar Glaser, Tamara Best, Thilo Kohler, Thomas Robinson, Tianhe Li, Tianjun Zhang, Tim Matthews, Timothy Chou, Tzook Shaked, Varun Vontimitta, Victoria Ajayi, Victoria Montanez, Vijai Mohan, Vinay~Satish Kumar, Vishal Mangla, Vítor Albiero, Vlad Ionescu, Vlad Poenaru, Vlad~Tiberiu Mihailescu, Vladimir Ivanov, Wei Li, Wenchen Wang, Wenwen Jiang, Wes Bouaziz, Will Constable, Xiaocheng Tang, Xiaofang Wang, Xiaojian Wu, Xiaolan Wang, Xide Xia, Xilun Wu, Xinbo Gao, Yanjun Chen, Ye~Hu, Ye~Jia, Ye~Qi, Yenda Li, Yilin Zhang, Ying Zhang, Yossi Adi, Youngjin Nam, Yu, Wang, Yuchen Hao, Yundi Qian, Yuzi He, Zach Rait, Zachary DeVito, Zef Rosnbrick, Zhaoduo Wen, Zhenyu Yang, and Zhiwei Zhao.
\newblock The llama 3 herd of models, 2024.
\newblock URL \url{https://arxiv.org/abs/2407.21783}.

\bibitem[Gao et~al.(2024)Gao, Song, Yang, Cai, Miao, Dong, Li, Ma, Chen, Xu, Tang, Wang, Zan, Quan, Zhang, Sha, Zhang, Ren, Liu, and Chang]{gao2024omnimathuniversalolympiadlevel}
Bofei Gao, Feifan Song, Zhe Yang, Zefan Cai, Yibo Miao, Qingxiu Dong, Lei Li, Chenghao Ma, Liang Chen, Runxin Xu, Zhengyang Tang, Benyou Wang, Daoguang Zan, Shanghaoran Quan, Ge~Zhang, Lei Sha, Yichang Zhang, Xuancheng Ren, Tianyu Liu, and Baobao Chang.
\newblock Omni-math: A universal olympiad level mathematic benchmark for large language models, 2024.
\newblock URL \url{https://arxiv.org/abs/2410.07985}.

\bibitem[Grave et~al.(2018)Grave, Bojanowski, Gupta, Joulin, and Mikolov]{grave2018learning}
Edouard Grave, Piotr Bojanowski, Prakhar Gupta, Armand Joulin, and Tomas Mikolov.
\newblock Learning word vectors for 157 languages.
\newblock In \emph{Proceedings of the International Conference on Language Resources and Evaluation (LREC 2018)}, 2018.

\bibitem[Guan et~al.(2025)Guan, Zhang, Liu, Shang, Sun, Zhu, Yang, and Yang]{guan2025rstarmathsmallllmsmaster}
Xinyu Guan, Li~Lyna Zhang, Yifei Liu, Ning Shang, Youran Sun, Yi~Zhu, Fan Yang, and Mao Yang.
\newblock rstar-math: Small llms can master math reasoning with self-evolved deep thinking, 2025.
\newblock URL \url{https://arxiv.org/abs/2501.04519}.

\bibitem[Havrilla et~al.(2024{\natexlab{a}})Havrilla, Dai, O'Mahony, Oostermeijer, Zisler, Albalak, Milo, Raparthy, Gandhi, Abbasi, Phung, Iyer, Mahan, Blagden, Gureja, Hamdy, Li, Paolini, Ammanamanchi, and Meyerson]{havrilla2024surveyingeffectsqualitydiversity}
Alex Havrilla, Andrew Dai, Laura O'Mahony, Koen Oostermeijer, Vera Zisler, Alon Albalak, Fabrizio Milo, Sharath~Chandra Raparthy, Kanishk Gandhi, Baber Abbasi, Duy Phung, Maia Iyer, Dakota Mahan, Chase Blagden, Srishti Gureja, Mohammed Hamdy, Wen-Ding Li, Giovanni Paolini, Pawan~Sasanka Ammanamanchi, and Elliot Meyerson.
\newblock Surveying the effects of quality, diversity, and complexity in synthetic data from large language models, 2024{\natexlab{a}}.
\newblock URL \url{https://arxiv.org/abs/2412.02980}.

\bibitem[Havrilla et~al.(2024{\natexlab{b}})Havrilla, Raparthy, Nalmpantis, Dwivedi-Yu, Zhuravinskyi, Hambro, and Raileanu]{havrilla2024glorewhenwhereimprove}
Alex Havrilla, Sharath Raparthy, Christoforus Nalmpantis, Jane Dwivedi-Yu, Maksym Zhuravinskyi, Eric Hambro, and Roberta Raileanu.
\newblock Glore: When, where, and how to improve llm reasoning via global and local refinements, 2024{\natexlab{b}}.
\newblock URL \url{https://arxiv.org/abs/2402.10963}.

\bibitem[Hendrycks et~al.(2021)Hendrycks, Burns, Kadavath, Arora, Basart, Tang, Song, and Steinhardt]{hendrycks2021measuringmathematicalproblemsolving}
Dan Hendrycks, Collin Burns, Saurav Kadavath, Akul Arora, Steven Basart, Eric Tang, Dawn Song, and Jacob Steinhardt.
\newblock Measuring mathematical problem solving with the math dataset, 2021.
\newblock URL \url{https://arxiv.org/abs/2103.03874}.

\bibitem[Hou et~al.(2025)Hou, Lv, Lu, Zhang, Li, Yao, Li, Tang, and Dong]{hou2025advancinglanguagemodelreasoning}
Zhenyu Hou, Xin Lv, Rui Lu, Jiajie Zhang, Yujiang Li, Zijun Yao, Juanzi Li, Jie Tang, and Yuxiao Dong.
\newblock Advancing language model reasoning through reinforcement learning and inference scaling, 2025.
\newblock URL \url{https://arxiv.org/abs/2501.11651}.

\bibitem[Jiang et~al.(2021)Jiang, Grefenstette, and Rocktäschel]{jiang2021prioritizedlevelreplay}
Minqi Jiang, Edward Grefenstette, and Tim Rocktäschel.
\newblock Prioritized level replay, 2021.
\newblock URL \url{https://arxiv.org/abs/2010.03934}.

\bibitem[Joulin et~al.(2016{\natexlab{a}})Joulin, Grave, Bojanowski, Douze, J{'e}gou, and Mikolov]{joulin2016fasttext}
Armand Joulin, Edouard Grave, Piotr Bojanowski, Matthijs Douze, H{'e}rve J{'e}gou, and Tomas Mikolov.
\newblock Fasttext.zip: Compressing text classification models.
\newblock \emph{arXiv preprint arXiv:1612.03651}, 2016{\natexlab{a}}.

\bibitem[Joulin et~al.(2016{\natexlab{b}})Joulin, Grave, Bojanowski, and Mikolov]{joulin2016bag}
Armand Joulin, Edouard Grave, Piotr Bojanowski, and Tomas Mikolov.
\newblock Bag of tricks for efficient text classification.
\newblock \emph{arXiv preprint arXiv:1607.01759}, 2016{\natexlab{b}}.

\bibitem[Kim et~al.(2025)Kim, Wu, Lee, and Suzuki]{kim2025metastabledynamicschainofthoughtreasoning}
Juno Kim, Denny Wu, Jason Lee, and Taiji Suzuki.
\newblock Metastable dynamics of chain-of-thought reasoning: Provable benefits of search, rl and distillation, 2025.
\newblock URL \url{https://arxiv.org/abs/2502.01694}.

\bibitem[Labs(2025)]{bespoke_stratos}
Bespoke Labs.
\newblock Bespoke-stratos: The unreasonable effectiveness of reasoning distillation, 2025.
\newblock URL \url{https://hf.co/bespokelabs/Bespoke-Stratos-32B}.
\newblock Accessed: 2025-01-22.

\bibitem[Lambert et~al.(2024)Lambert, Morrison, Pyatkin, Huang, Ivison, Brahman, Miranda, Liu, Dziri, Lyu, Gu, Malik, Graf, Hwang, Yang, Bras, Tafjord, Wilhelm, Soldaini, Smith, Wang, Dasigi, and Hajishirzi]{lambert2024tulu3pushingfrontiers}
Nathan Lambert, Jacob Morrison, Valentina Pyatkin, Shengyi Huang, Hamish Ivison, Faeze Brahman, Lester James~V. Miranda, Alisa Liu, Nouha Dziri, Shane Lyu, Yuling Gu, Saumya Malik, Victoria Graf, Jena~D. Hwang, Jiangjiang Yang, Ronan~Le Bras, Oyvind Tafjord, Chris Wilhelm, Luca Soldaini, Noah~A. Smith, Yizhong Wang, Pradeep Dasigi, and Hannaneh Hajishirzi.
\newblock Tulu 3: Pushing frontiers in open language model post-training, 2024.
\newblock URL \url{https://arxiv.org/abs/2411.15124}.

\bibitem[Li et~al.(2025)Li, Cao, Griggs, Liu, Mo, Patil, Zaharia, Gonzalez, and Stoica]{li2025llmseasilylearnreason}
Dacheng Li, Shiyi Cao, Tyler Griggs, Shu Liu, Xiangxi Mo, Shishir~G. Patil, Matei Zaharia, Joseph~E. Gonzalez, and Ion Stoica.
\newblock Llms can easily learn to reason from demonstrations structure, not content, is what matters!, 2025.
\newblock URL \url{https://arxiv.org/abs/2502.07374}.

\bibitem[Li et~al.(2024{\natexlab{a}})Li, Fang, Smyrnis, Ivgi, Jordan, Gadre, Bansal, Guha, Keh, Arora, Garg, Xin, Muennighoff, Heckel, Mercat, Chen, Gururangan, Wortsman, Albalak, Bitton, Nezhurina, Abbas, Hsieh, Ghosh, Gardner, Kilian, Zhang, Shao, Pratt, Sanyal, Ilharco, Daras, Marathe, Gokaslan, Zhang, Chandu, Nguyen, Vasiljevic, Kakade, Song, Sanghavi, Faghri, Oh, Zettlemoyer, Lo, El-Nouby, Pouransari, Toshev, Wang, Groeneveld, Soldaini, Koh, Jitsev, Kollar, Dimakis, Carmon, Dave, Schmidt, and Shankar]{li2024datacomplmsearchgenerationtraining}
Jeffrey Li, Alex Fang, Georgios Smyrnis, Maor Ivgi, Matt Jordan, Samir Gadre, Hritik Bansal, Etash Guha, Sedrick Keh, Kushal Arora, Saurabh Garg, Rui Xin, Niklas Muennighoff, Reinhard Heckel, Jean Mercat, Mayee Chen, Suchin Gururangan, Mitchell Wortsman, Alon Albalak, Yonatan Bitton, Marianna Nezhurina, Amro Abbas, Cheng-Yu Hsieh, Dhruba Ghosh, Josh Gardner, Maciej Kilian, Hanlin Zhang, Rulin Shao, Sarah Pratt, Sunny Sanyal, Gabriel Ilharco, Giannis Daras, Kalyani Marathe, Aaron Gokaslan, Jieyu Zhang, Khyathi Chandu, Thao Nguyen, Igor Vasiljevic, Sham Kakade, Shuran Song, Sujay Sanghavi, Fartash Faghri, Sewoong Oh, Luke Zettlemoyer, Kyle Lo, Alaaeldin El-Nouby, Hadi Pouransari, Alexander Toshev, Stephanie Wang, Dirk Groeneveld, Luca Soldaini, Pang~Wei Koh, Jenia Jitsev, Thomas Kollar, Alexandros~G. Dimakis, Yair Carmon, Achal Dave, Ludwig Schmidt, and Vaishaal Shankar.
\newblock Datacomp-lm: In search of the next generation of training sets for language models, 2024{\natexlab{a}}.
\newblock URL \url{https://arxiv.org/abs/2406.11794}.

\bibitem[Li et~al.(2024{\natexlab{b}})Li, Beeching, Tunstall, Lipkin, Soletskyi, Huang, Rasul, Yu, Jiang, Shen, Qin, Dong, Zhou, Fleureau, Lample, and Polu]{numina_math_datasets}
Jia Li, Edward Beeching, Lewis Tunstall, Ben Lipkin, Roman Soletskyi, Shengyi~Costa Huang, Kashif Rasul, Longhui Yu, Albert Jiang, Ziju Shen, Zihan Qin, Bin Dong, Li~Zhou, Yann Fleureau, Guillaume Lample, and Stanislas Polu.
\newblock Numinamath.
\newblock \url{[https://huggingface.co/AI-MO/NuminaMath-CoT](https://github.com/project-numina/aimo-progress-prize/blob/main/report/numina_dataset.pdf)}, 2024{\natexlab{b}}.

\bibitem[Lightman et~al.(2023{\natexlab{a}})Lightman, Kosaraju, Burda, Edwards, Baker, Lee, Leike, Schulman, Sutskever, and Cobbe]{lightman2023let}
Hunter Lightman, Vineet Kosaraju, Yura Burda, Harri Edwards, Bowen Baker, Teddy Lee, Jan Leike, John Schulman, Ilya Sutskever, and Karl Cobbe.
\newblock Let's verify step by step.
\newblock \emph{arXiv preprint arXiv:2305.20050}, 2023{\natexlab{a}}.

\bibitem[Lightman et~al.(2023{\natexlab{b}})Lightman, Kosaraju, Burda, Edwards, Baker, Lee, Leike, Schulman, Sutskever, and Cobbe]{lightman2023letsverifystepstep}
Hunter Lightman, Vineet Kosaraju, Yura Burda, Harri Edwards, Bowen Baker, Teddy Lee, Jan Leike, John Schulman, Ilya Sutskever, and Karl Cobbe.
\newblock Let's verify step by step, 2023{\natexlab{b}}.
\newblock URL \url{https://arxiv.org/abs/2305.20050}.

\bibitem[Liu et~al.(2024)Liu, Chen, Shoeybi, Catanzaro, and Ping]{liu2024acemathadvancingfrontiermath}
Zihan Liu, Yang Chen, Mohammad Shoeybi, Bryan Catanzaro, and Wei Ping.
\newblock Acemath: Advancing frontier math reasoning with post-training and reward modeling, 2024.
\newblock URL \url{https://arxiv.org/abs/2412.15084}.

\bibitem[Mahan et~al.(2024)Mahan, Phung, Rafailov, Blagden, Lile, Castricato, Fränken, Finn, and Albalak]{mahan2024generativerewardmodels}
Dakota Mahan, Duy~Van Phung, Rafael Rafailov, Chase Blagden, Nathan Lile, Louis Castricato, Jan-Philipp Fränken, Chelsea Finn, and Alon Albalak.
\newblock Generative reward models, 2024.
\newblock URL \url{https://arxiv.org/abs/2410.12832}.

\bibitem[Mirzadeh et~al.(2024)Mirzadeh, Alizadeh, Shahrokhi, Tuzel, Bengio, and Farajtabar]{mirzadeh2024gsmsymbolicunderstandinglimitationsmathematical}
Iman Mirzadeh, Keivan Alizadeh, Hooman Shahrokhi, Oncel Tuzel, Samy Bengio, and Mehrdad Farajtabar.
\newblock Gsm-symbolic: Understanding the limitations of mathematical reasoning in large language models, 2024.
\newblock URL \url{https://arxiv.org/abs/2410.05229}.

\bibitem[Mitra et~al.(2024)Mitra, Khanpour, Rosset, and Awadallah]{mitra2024orcamathunlockingpotentialslms}
Arindam Mitra, Hamed Khanpour, Corby Rosset, and Ahmed Awadallah.
\newblock Orca-math: Unlocking the potential of slms in grade school math, 2024.
\newblock URL \url{https://arxiv.org/abs/2402.14830}.

\bibitem[Muennighoff et~al.(2025)Muennighoff, Yang, Shi, Li, Fei-Fei, Hajishirzi, Zettlemoyer, Liang, Candès, and Hashimoto]{muennighoff2025s1simpletesttimescaling}
Niklas Muennighoff, Zitong Yang, Weijia Shi, Xiang~Lisa Li, Li~Fei-Fei, Hannaneh Hajishirzi, Luke Zettlemoyer, Percy Liang, Emmanuel Candès, and Tatsunori Hashimoto.
\newblock s1: Simple test-time scaling, 2025.
\newblock URL \url{https://arxiv.org/abs/2501.19393}.

\bibitem[of~America(2024)]{aime}
Mathematical~Association of~America.
\newblock Aime, February 2024.
\newblock URL \url{https://artofproblemsolving.com/wiki/index.php/AIME_Problems_and_Solutions/}.

\bibitem[OpenAI et~al.(2024)OpenAI, :, Jaech, Kalai, Lerer, Richardson, El-Kishky, Low, Helyar, Madry, Beutel, Carney, Iftimie, Karpenko, Passos, Neitz, Prokofiev, Wei, Tam, Bennett, Kumar, Saraiva, Vallone, Duberstein, Kondrich, Mishchenko, Applebaum, Jiang, Nair, Zoph, Ghorbani, Rossen, Sokolowsky, Barak, McGrew, Minaiev, Hao, Baker, Houghton, McKinzie, Eastman, Lugaresi, Bassin, Hudson, Li, de~Bourcy, Voss, Shen, Zhang, Koch, Orsinger, Hesse, Fischer, Chan, Roberts, Kappler, Levy, Selsam, Dohan, Farhi, Mely, Robinson, Tsipras, Li, Oprica, Freeman, Zhang, Wong, Proehl, Cheung, Mitchell, Wallace, Ritter, Mays, Wang, Such, Raso, Leoni, Tsimpourlas, Song, von Lohmann, Sulit, Salmon, Parascandolo, Chabot, Zhao, Brockman, Leclerc, Salman, Bao, Sheng, Andrin, Bagherinezhad, Ren, Lightman, Chung, Kivlichan, O'Connell, Osband, Gilaberte, Akkaya, Kostrikov, Sutskever, Kofman, Pachocki, Lennon, Wei, Harb, Twore, Feng, Yu, Weng, Tang, Yu, Candela, Palermo, Parish, Heidecke, Hallman, Rizzo, Gordon, Uesato, Ward,
  Huizinga, Wang, Chen, Xiao, Singhal, Nguyen, Cobbe, Shi, Wood, Rimbach, Gu-Lemberg, Liu, Lu, Stone, Yu, Ahmad, Yang, Liu, Maksin, Ho, Fedus, Weng, Li, McCallum, Held, Kuhn, Kondraciuk, Kaiser, Metz, Boyd, Trebacz, Joglekar, Chen, Tintor, Meyer, Jones, Kaufer, Schwarzer, Shah, Yatbaz, Guan, Xu, Yan, Glaese, Chen, Lampe, Malek, Wang, Fradin, McClay, Pavlov, Wang, Wang, Murati, Bavarian, Rohaninejad, McAleese, Chowdhury, Chowdhury, Ryder, Tezak, Brown, Nachum, Boiko, Murk, Watkins, Chao, Ashbourne, Izmailov, Zhokhov, Dias, Arora, Lin, Lopes, Gaon, Miyara, Leike, Hwang, Garg, Brown, James, Shu, Cheu, Greene, Jain, Altman, Toizer, Toyer, Miserendino, Agarwal, Hernandez, Baker, McKinney, Yan, Zhao, Hu, Santurkar, Chaudhuri, Zhang, Fu, Papay, Lin, Balaji, Sanjeev, Sidor, Broda, Clark, Wang, Gordon, Sanders, Patwardhan, Sottiaux, Degry, Dimson, Zheng, Garipov, Stasi, Bansal, Creech, Peterson, Eloundou, Qi, Kosaraju, Monaco, Pong, Fomenko, Zheng, Zhou, McCabe, Zaremba, Dubois, Lu, Chen, Cha, Bai, He, Zhang, Wang,
  Shao, and Li]{openai2024openaio1card}
OpenAI, :, Aaron Jaech, Adam Kalai, Adam Lerer, Adam Richardson, Ahmed El-Kishky, Aiden Low, Alec Helyar, Aleksander Madry, Alex Beutel, Alex Carney, Alex Iftimie, Alex Karpenko, Alex~Tachard Passos, Alexander Neitz, Alexander Prokofiev, Alexander Wei, Allison Tam, Ally Bennett, Ananya Kumar, Andre Saraiva, Andrea Vallone, Andrew Duberstein, Andrew Kondrich, Andrey Mishchenko, Andy Applebaum, Angela Jiang, Ashvin Nair, Barret Zoph, Behrooz Ghorbani, Ben Rossen, Benjamin Sokolowsky, Boaz Barak, Bob McGrew, Borys Minaiev, Botao Hao, Bowen Baker, Brandon Houghton, Brandon McKinzie, Brydon Eastman, Camillo Lugaresi, Cary Bassin, Cary Hudson, Chak~Ming Li, Charles de~Bourcy, Chelsea Voss, Chen Shen, Chong Zhang, Chris Koch, Chris Orsinger, Christopher Hesse, Claudia Fischer, Clive Chan, Dan Roberts, Daniel Kappler, Daniel Levy, Daniel Selsam, David Dohan, David Farhi, David Mely, David Robinson, Dimitris Tsipras, Doug Li, Dragos Oprica, Eben Freeman, Eddie Zhang, Edmund Wong, Elizabeth Proehl, Enoch Cheung, Eric
  Mitchell, Eric Wallace, Erik Ritter, Evan Mays, Fan Wang, Felipe~Petroski Such, Filippo Raso, Florencia Leoni, Foivos Tsimpourlas, Francis Song, Fred von Lohmann, Freddie Sulit, Geoff Salmon, Giambattista Parascandolo, Gildas Chabot, Grace Zhao, Greg Brockman, Guillaume Leclerc, Hadi Salman, Haiming Bao, Hao Sheng, Hart Andrin, Hessam Bagherinezhad, Hongyu Ren, Hunter Lightman, Hyung~Won Chung, Ian Kivlichan, Ian O'Connell, Ian Osband, Ignasi~Clavera Gilaberte, Ilge Akkaya, Ilya Kostrikov, Ilya Sutskever, Irina Kofman, Jakub Pachocki, James Lennon, Jason Wei, Jean Harb, Jerry Twore, Jiacheng Feng, Jiahui Yu, Jiayi Weng, Jie Tang, Jieqi Yu, Joaquin~Quiñonero Candela, Joe Palermo, Joel Parish, Johannes Heidecke, John Hallman, John Rizzo, Jonathan Gordon, Jonathan Uesato, Jonathan Ward, Joost Huizinga, Julie Wang, Kai Chen, Kai Xiao, Karan Singhal, Karina Nguyen, Karl Cobbe, Katy Shi, Kayla Wood, Kendra Rimbach, Keren Gu-Lemberg, Kevin Liu, Kevin Lu, Kevin Stone, Kevin Yu, Lama Ahmad, Lauren Yang, Leo Liu,
  Leon Maksin, Leyton Ho, Liam Fedus, Lilian Weng, Linden Li, Lindsay McCallum, Lindsey Held, Lorenz Kuhn, Lukas Kondraciuk, Lukasz Kaiser, Luke Metz, Madelaine Boyd, Maja Trebacz, Manas Joglekar, Mark Chen, Marko Tintor, Mason Meyer, Matt Jones, Matt Kaufer, Max Schwarzer, Meghan Shah, Mehmet Yatbaz, Melody~Y. Guan, Mengyuan Xu, Mengyuan Yan, Mia Glaese, Mianna Chen, Michael Lampe, Michael Malek, Michele Wang, Michelle Fradin, Mike McClay, Mikhail Pavlov, Miles Wang, Mingxuan Wang, Mira Murati, Mo~Bavarian, Mostafa Rohaninejad, Nat McAleese, Neil Chowdhury, Neil Chowdhury, Nick Ryder, Nikolas Tezak, Noam Brown, Ofir Nachum, Oleg Boiko, Oleg Murk, Olivia Watkins, Patrick Chao, Paul Ashbourne, Pavel Izmailov, Peter Zhokhov, Rachel Dias, Rahul Arora, Randall Lin, Rapha~Gontijo Lopes, Raz Gaon, Reah Miyara, Reimar Leike, Renny Hwang, Rhythm Garg, Robin Brown, Roshan James, Rui Shu, Ryan Cheu, Ryan Greene, Saachi Jain, Sam Altman, Sam Toizer, Sam Toyer, Samuel Miserendino, Sandhini Agarwal, Santiago Hernandez,
  Sasha Baker, Scott McKinney, Scottie Yan, Shengjia Zhao, Shengli Hu, Shibani Santurkar, Shraman~Ray Chaudhuri, Shuyuan Zhang, Siyuan Fu, Spencer Papay, Steph Lin, Suchir Balaji, Suvansh Sanjeev, Szymon Sidor, Tal Broda, Aidan Clark, Tao Wang, Taylor Gordon, Ted Sanders, Tejal Patwardhan, Thibault Sottiaux, Thomas Degry, Thomas Dimson, Tianhao Zheng, Timur Garipov, Tom Stasi, Trapit Bansal, Trevor Creech, Troy Peterson, Tyna Eloundou, Valerie Qi, Vineet Kosaraju, Vinnie Monaco, Vitchyr Pong, Vlad Fomenko, Weiyi Zheng, Wenda Zhou, Wes McCabe, Wojciech Zaremba, Yann Dubois, Yinghai Lu, Yining Chen, Young Cha, Yu~Bai, Yuchen He, Yuchen Zhang, Yunyun Wang, Zheng Shao, and Zhuohan Li.
\newblock Openai o1 system card, 2024.
\newblock URL \url{https://arxiv.org/abs/2412.16720}.

\bibitem[OpenAI et~al.(2025)OpenAI, :, El-Kishky, Wei, Saraiva, Minaev, Selsam, Dohan, Song, Lightman, Clavera, Pachocki, Tworek, Kuhn, Kaiser, Chen, Schwarzer, Rohaninejad, McAleese, o3~contributors, Mürk, Garg, Shu, Sidor, Kosaraju, and Zhou]{openai2025competitiveprogramminglargereasoning}
OpenAI, :, Ahmed El-Kishky, Alexander Wei, Andre Saraiva, Borys Minaev, Daniel Selsam, David Dohan, Francis Song, Hunter Lightman, Ignasi Clavera, Jakub Pachocki, Jerry Tworek, Lorenz Kuhn, Lukasz Kaiser, Mark Chen, Max Schwarzer, Mostafa Rohaninejad, Nat McAleese, o3~contributors, Oleg Mürk, Rhythm Garg, Rui Shu, Szymon Sidor, Vineet Kosaraju, and Wenda Zhou.
\newblock Competitive programming with large reasoning models, 2025.
\newblock URL \url{https://arxiv.org/abs/2502.06807}.

\bibitem[Schaeffer et~al.(2023)Schaeffer, Miranda, and Koyejo]{schaeffer2023emergentabilitieslargelanguage}
Rylan Schaeffer, Brando Miranda, and Sanmi Koyejo.
\newblock Are emergent abilities of large language models a mirage?, 2023.
\newblock URL \url{https://arxiv.org/abs/2304.15004}.

\bibitem[Setlur et~al.(2025)Setlur, Rajaraman, Levine, and Kumar]{setlur2025scalingtesttimecomputeverification}
Amrith Setlur, Nived Rajaraman, Sergey Levine, and Aviral Kumar.
\newblock Scaling test-time compute without verification or rl is suboptimal, 2025.
\newblock URL \url{https://arxiv.org/abs/2502.12118}.

\bibitem[Soldaini et~al.(2024)Soldaini, Kinney, Bhagia, Schwenk, Atkinson, Authur, Bogin, Chandu, Dumas, Elazar, Hofmann, Jha, Kumar, Lucy, Lyu, Lambert, Magnusson, Morrison, Muennighoff, Naik, Nam, Peters, Ravichander, Richardson, Shen, Strubell, Subramani, Tafjord, Walsh, Zettlemoyer, Smith, Hajishirzi, Beltagy, Groeneveld, Dodge, and Lo]{soldaini2024dolmaopencorpustrillion}
Luca Soldaini, Rodney Kinney, Akshita Bhagia, Dustin Schwenk, David Atkinson, Russell Authur, Ben Bogin, Khyathi Chandu, Jennifer Dumas, Yanai Elazar, Valentin Hofmann, Ananya~Harsh Jha, Sachin Kumar, Li~Lucy, Xinxi Lyu, Nathan Lambert, Ian Magnusson, Jacob Morrison, Niklas Muennighoff, Aakanksha Naik, Crystal Nam, Matthew~E. Peters, Abhilasha Ravichander, Kyle Richardson, Zejiang Shen, Emma Strubell, Nishant Subramani, Oyvind Tafjord, Pete Walsh, Luke Zettlemoyer, Noah~A. Smith, Hannaneh Hajishirzi, Iz~Beltagy, Dirk Groeneveld, Jesse Dodge, and Kyle Lo.
\newblock Dolma: an open corpus of three trillion tokens for language model pretraining research, 2024.
\newblock URL \url{https://arxiv.org/abs/2402.00159}.

\bibitem[Team et~al.(2025)Team, Du, Gao, Xing, Jiang, Chen, Li, Xiao, Du, Liao, Tang, Wang, Zhang, Yuan, Lu, Tang, Sung, Wei, Lai, Guo, Zhu, Ding, Hu, Yang, Zhang, Yao, Zhao, Lu, Li, Yu, Gao, Zheng, Yuan, Chen, Guo, Su, Wang, Zhao, Zhang, Liu, Yan, Wu, Shi, Ye, Yu, Dong, Zhang, Ma, Pan, Gong, Liu, Ma, Wei, Cao, Huang, Jiang, Gao, Xiong, He, Huang, Wu, He, Wei, Jia, Wu, Xu, Zu, Zhou, Pan, Charles, Li, Hu, Liu, Chen, Wang, Liu, Qin, Liu, Yang, Bao, Du, Wu, Wang, Zhou, Wang, Li, Zhu, Zhang, Wang, Yang, Huang, Huang, Xu, and Yang]{kimiteam2025kimik15scalingreinforcement}
Kimi Team, Angang Du, Bofei Gao, Bowei Xing, Changjiu Jiang, Cheng Chen, Cheng Li, Chenjun Xiao, Chenzhuang Du, Chonghua Liao, Chuning Tang, Congcong Wang, Dehao Zhang, Enming Yuan, Enzhe Lu, Fengxiang Tang, Flood Sung, Guangda Wei, Guokun Lai, Haiqing Guo, Han Zhu, Hao Ding, Hao Hu, Hao Yang, Hao Zhang, Haotian Yao, Haotian Zhao, Haoyu Lu, Haoze Li, Haozhen Yu, Hongcheng Gao, Huabin Zheng, Huan Yuan, Jia Chen, Jianhang Guo, Jianlin Su, Jianzhou Wang, Jie Zhao, Jin Zhang, Jingyuan Liu, Junjie Yan, Junyan Wu, Lidong Shi, Ling Ye, Longhui Yu, Mengnan Dong, Neo Zhang, Ningchen Ma, Qiwei Pan, Qucheng Gong, Shaowei Liu, Shengling Ma, Shupeng Wei, Sihan Cao, Siying Huang, Tao Jiang, Weihao Gao, Weimin Xiong, Weiran He, Weixiao Huang, Wenhao Wu, Wenyang He, Xianghui Wei, Xianqing Jia, Xingzhe Wu, Xinran Xu, Xinxing Zu, Xinyu Zhou, Xuehai Pan, Y.~Charles, Yang Li, Yangyang Hu, Yangyang Liu, Yanru Chen, Yejie Wang, Yibo Liu, Yidao Qin, Yifeng Liu, Ying Yang, Yiping Bao, Yulun Du, Yuxin Wu, Yuzhi Wang, Zaida Zhou,
  Zhaoji Wang, Zhaowei Li, Zhen Zhu, Zheng Zhang, Zhexu Wang, Zhilin Yang, Zhiqi Huang, Zihao Huang, Ziyao Xu, and Zonghan Yang.
\newblock Kimi k1.5: Scaling reinforcement learning with llms, 2025.
\newblock URL \url{https://arxiv.org/abs/2501.12599}.

\bibitem[Wang et~al.(2024)Wang, Wu, Song, Mittal, and Jia]{NEURIPS2024_ed165f2f}
Jiachen~(Tianhao) Wang, Tong Wu, Dawn Song, Prateek Mittal, and Ruoxi Jia.
\newblock Greats: Online selection of high-quality data for llm training in every iteration.
\newblock In A.~Globerson, L.~Mackey, D.~Belgrave, A.~Fan, U.~Paquet, J.~Tomczak, and C.~Zhang, editors, \emph{Advances in Neural Information Processing Systems}, volume~37, pages 131197--131223. Curran Associates, Inc., 2024.
\newblock URL \url{https://proceedings.neurips.cc/paper_files/paper/2024/file/ed165f2ff227cf36c7e3ef88957dadd9-Paper-Conference.pdf}.

\bibitem[Wang et~al.(2025)Wang, Yang, Wang, and Wei]{wang2025examiningfalsepositivesinference}
Yu~Wang, Nan Yang, Liang Wang, and Furu Wei.
\newblock Examining false positives under inference scaling for mathematical reasoning, 2025.
\newblock URL \url{https://arxiv.org/abs/2502.06217}.

\bibitem[Wei et~al.(2023)Wei, Wang, Schuurmans, Bosma, Ichter, Xia, Chi, Le, and Zhou]{wei2023chainofthoughtpromptingelicitsreasoning}
Jason Wei, Xuezhi Wang, Dale Schuurmans, Maarten Bosma, Brian Ichter, Fei Xia, Ed~Chi, Quoc Le, and Denny Zhou.
\newblock Chain-of-thought prompting elicits reasoning in large language models, 2023.
\newblock URL \url{https://arxiv.org/abs/2201.11903}.

\bibitem[Xiang et~al.(2025)Xiang, Snell, Gandhi, Albalak, Singh, Blagden, Phung, Rafailov, Lile, Mahan, Castricato, Franken, Haber, and Finn]{xiang20252reasoningllmslearning}
Violet Xiang, Charlie Snell, Kanishk Gandhi, Alon Albalak, Anikait Singh, Chase Blagden, Duy Phung, Rafael Rafailov, Nathan Lile, Dakota Mahan, Louis Castricato, Jan-Philipp Franken, Nick Haber, and Chelsea Finn.
\newblock Towards system 2 reasoning in llms: Learning how to think with meta chain-of-thought, 2025.
\newblock URL \url{https://arxiv.org/abs/2501.04682}.

\bibitem[Yeo et~al.(2025)Yeo, Tong, Niu, Neubig, and Yue]{yeo2025demystifyinglongchainofthoughtreasoning}
Edward Yeo, Yuxuan Tong, Morry Niu, Graham Neubig, and Xiang Yue.
\newblock Demystifying long chain-of-thought reasoning in llms, 2025.
\newblock URL \url{https://arxiv.org/abs/2502.03373}.

\bibitem[Yuan et~al.(2024)Yuan, Li, Chen, Cui, Ding, Zhang, Zhou, Liu, and Peng]{yuan2024freeprocessrewardsprocess}
Lifan Yuan, Wendi Li, Huayu Chen, Ganqu Cui, Ning Ding, Kaiyan Zhang, Bowen Zhou, Zhiyuan Liu, and Hao Peng.
\newblock Free process rewards without process labels, 2024.
\newblock URL \url{https://arxiv.org/abs/2412.01981}.

\bibitem[Yuan et~al.(2025)Yuan, Yu, Jiang, Padthe, Li, Wang, Kulikov, Cho, Tian, Weston, and Li]{yuan2025naturalreasoningreasoningwild28m}
Weizhe Yuan, Jane Yu, Song Jiang, Karthik Padthe, Yang Li, Dong Wang, Ilia Kulikov, Kyunghyun Cho, Yuandong Tian, Jason~E Weston, and Xian Li.
\newblock Naturalreasoning: Reasoning in the wild with 2.8m challenging questions, 2025.
\newblock URL \url{https://arxiv.org/abs/2502.13124}.

\bibitem[Yue et~al.(2024)Yue, Madaan, Moskovitz, Strouse, and Singh]{yue2024harp}
Albert~S. Yue, Lovish Madaan, Ted Moskovitz, DJ~Strouse, and Aaditya~K. Singh.
\newblock {HARP}: A challenging human-annotated math reasoning benchmark, 2024.
\newblock URL \url{https://github.com/aadityasingh/HARP}.

\bibitem[Zhang et~al.(2024)Zhang, Hosseini, Bansal, Kazemi, Kumar, and Agarwal]{zhang2024generativeverifiersrewardmodeling}
Lunjun Zhang, Arian Hosseini, Hritik Bansal, Mehran Kazemi, Aviral Kumar, and Rishabh Agarwal.
\newblock Generative verifiers: Reward modeling as next-token prediction, 2024.
\newblock URL \url{https://arxiv.org/abs/2408.15240}.

\end{thebibliography}
\bibliographystyle{plainnat}

\newpage
\appendix
\onecolumn

\section{Filter Statistics}
\begin{center}
    \begin{adjustbox}{angle=90}
\centering
\small
\begin{tabular}{|l|ccccccccccc|c|}
\toprule
 & \multicolumn{2}{c}{\textbf{Multiple Choice}} & \multicolumn{2}{c}{\textbf{True/False}} & \multicolumn{2}{c}{\textbf{Yes/No}} & \multicolumn{2}{c}{\textbf{Multi-Part}} & \multicolumn{2}{c}{\textbf{Proof}} & \textbf{Semantic} & \textbf{Total} \\
\textbf{Subset} &(R) & (L) & (R)&(L)&(R)&(L)&(R)&(L)&(R)&(L)& \textbf{Deduplication}&\\
\midrule
Orca-Math & 1174 & 111 & 5 & 0 & 12 & 0 & 16345 & 2298 & 2 & 5 & 8286 & 26,487 \\
cn\_k12 & 94888 & 94932 & 420 & 0 & 384 & 0 & 20358 & 19428 & 3969 & 5185 & 2098 & 123,809 \\
olympiads & 12123 & 9507 & 311 & 0 & 4615 & 0 & 13509 & 6981 & 23210 & 30315 & 1362 & 53,102 \\
MATH & 1046 & 386 & 0 & 0 & 0 & 3 & 2734 & 50 & 0 & 31 & 7694 & 9,714 \\
aops\_forum & 1588 & 1162 & 85 & 0 & 126 & 0 & 1508 & 1488 & 2282 & 4605 & 262 & 7,853 \\
GSM8k & 45 & 1 & 0 & 0 & 0 & 0 & 782 & 3 & 0 & 0 & 7664 & 7,999 \\
HARP & 92 & 215 & 2 & 0 & 7 & 0 & 253 & 44 & 0 & 2 & 155 & 695 \\
Omni-MATH & 125 & 29 & 7 & 122 & 191 & 169 & 283 & 212 & 3 & 207 & 378 & 1,182 \\
amc\_aime & 2955 & 2938 & 0 & 0 & 22 & 0 & 1018 & 7 & 36 & 63 & 53 & 3,066 \\
\textbf{Total} & 114,036 & 109,281 & 830 & 122 & 5,357 & 172 & 49,932 & 30,511 & 29,502 & 40,413 & 27,952 & 233,907 \\
\bottomrule
\end{tabular}
\label{app:filter_result_expanded}
\end{adjustbox}
\end{center}
\textbf{Statistics of the individual filters used to create \bigmath.} (R) columns are regular expression filters, while (L) columns are the result of filtering with Llama-3.1-70B.

\clearpage

\section{Domains By Source}
\begin{center}
    \begin{adjustbox}{angle=90}
    \centering
    \tiny
\begin{tabular}{|l|rrrrrrrrrrrrrr|}
\toprule
Domain & Abstract & Algebra & Calculus & Differential  & Discrete & Geometry & Intermediate & Linear & Math Word & Number & Other & Prealgebra & Precalculus & Statistics \\
Source & Algebra &  &  & Equations & Mathematics &  & Algebra & Algebra & Problems & Theory &  &  &  &  \\
\midrule
amc\_aime & 0 & 12 & 0 & 0 & 12 & 18 & 2 & 0 & 14 & 8 & 3 & 0 & 0 & 10 \\
aops\_forum  & 42 & 1263 & 212 & 3 & 848 & 1218 & 27 & 19 & 495 & 1065 & 174 & 4 & 39 & 458 \\
\bigmathmc & 17 & 14362 & 2725 & 2 & 1499 & 8207 & 3971 & 1600 & 4534 & 1219 & 939 & 1059 & 2629 & 5186 \\
cn\_k12 & 17 & 23282 & 6169 & 2 & 1757 & 12813 & 4324 & 1563 & 3033 & 774 & 1039 & 970 & 5022 & 4998 \\
GSM8k & 0 & 17 & 0 & 0 & 0 & 12 & 0 & 0 & 1810 & 0 & 0 & 11 & 0 & 15 \\
HARP & 1 & 677 & 21 & 0 & 123 & 690 & 101 & 4 & 716 & 250 & 36 & 50 & 28 & 340 \\
MATH & 12 & 2752 & 175 & 0 & 102 & 1247 & 532 & 282 & 992 & 1145 & 113 & 258 & 431 & 1018 \\
olympiads & 59 & 6458 & 1603 & 101 & 3347 & 9117 & 456 & 156 & 5744 & 2859 & 780 & 116 & 1029 & 2324 \\
Omni-MATH & 11 & 547 & 59 & 0 & 334 & 479 & 38 & 7 & 259 & 355 & 78 & 42 & 24 & 301 \\
Orca-Math  & 0 & 4689 & 29 & 1 & 111 & 3102 & 241 & 3 & 66563 & 2285 & 333 & 3091 & 11 & 2843 \\
\bottomrule
\end{tabular}
\label{tab:domain_by_source}
\end{adjustbox}
\end{center}
\textbf{Breakdown of problems by source and domains (as defined by~\citet{gao2024omnimathuniversalolympiadlevel}).}

\clearpage

\section{Reformulation Examples}
\label{app:reformulation_examples}

\begin{figure}
    \centering
    \begin{text_full}{Successful Reformulation}
    \textcolor{blue}{\textbf{Original Problem:}} What is the sum of the prime factors of 2010?\\(A) 67\\ (B) 75\\ (C) 77\\ (D) 201\\ (E) 210\\
    \textcolor{blue}{\textbf{Key Information Extraction:}}\\
\textbf{core\_mathematical\_concept}: "Number theory - prime factorization and sum of prime factors"\\ \textbf{key\_information\_extraction}: ["Find the sum of prime factors of 2010", "Prime factorization required"]\\
\textbf{problem\_structure\_analysis}: "Direct calculation problem"\\
\textbf{multiple\_choice\_removal\_strategy}: ["Remove all answer choices", "Ask for direct calculation of sum of prime factors"]\\
\textbf{rephrasing\_approach}: ["Request the sum of prime factors directly", "No comparative aspect needed"]\\
\textbf{problem\_integrity\_preservation}: ["Maintain the original number (2010)", "Require prime factorization"]\\
\textbf{answer\_format\_specification}: ["Answer should be a single integer", "No units or decimal places required"]\\
\textbf{is\_multiple\_choice}: true\\
    \textcolor{blue}{\textbf{Reformulation:}} Find the sum of the prime factors of 2010. Express your answer as a single integer.\\
    \textcolor{blue}{\textbf{Judgement:}} The reformulated problem is a direct and clear request to find the sum of the prime factors of 2010. It does not provide any multiple-choice options, nor does it limit the answer choices in any way. The problem requires the solver to find the prime factorization of 2010 and then sum the prime factors, which is a well-defined mathematical task.\\The original problem and the reformulated problem are mathematically equivalent, and the solution to the original problem is still applicable to the reformulated problem. The reformulated problem does not introduce any new mathematical concepts or difficulties, and it does not provide any additional information that would make the problem easier to solve.\\The answer format specification is clear and unambiguous, requiring the solver to express the answer as a single integer. This is a suitable format for the problem, as the sum of the prime factors is a well-defined integer value.\\Overall, the reformulated problem is a well-posed and mathematically sound problem that requires the solver to apply mathematical concepts and techniques to find the solution.\\
    \end{text_full}
    \begin{text_full}{Failed Reformulation}
    \textcolor{blue}{\textbf{Original Problem:}} Let $ x_{1}, x_{2}$ be distinct positive real numbers and $ a $ be a real number in the interval $ (0,1) $. Define $ y_{1} = \frac{x_{1}}{1+a} + \frac{a x_{2}}{1+a} $ and $ y_{2} = \frac{x_{2}}{1+a} + \frac{a x_{1}}{1+a} $. Determine the relationship between $ x_{1} x_{2} $ and $ y_{1} y_{2} $:\\(A) $ x_{1} \cdot x_{2} > y_{1} y_{2} $\\(B) $ x_{1} x_{2} = y_{1} y_{2} $\\(C) $ x_{1} x_{2} < y_{1} y_{2} $\\(D) Cannot be determined, it depends on the value of $ a $\\
    \textcolor{blue}{\textbf{Key Information Extraction:}}\\
\textbf{core\_mathematical\_concept}: "Inequalities involving real numbers and their products"\\
\textbf{key\_information\_extraction}: ["Distinct positive real numbers x1, x2", "Real number a in the interval (0,1)", "Expressions for y1 and y2 in terms of x1, x2, and a", "Need to compare the product x1*x2 with y1*y2"]\\
\textbf{problem\_structure\_analysis}: "Comparison problem requiring algebraic manipulation and inequality analysis"\\
\textbf{multiple\_choice\_removal\_strategy}: ["Remove all comparisons and answer choices", "Ask for direct determination of the relationship between x1*x2 and y1*y2"]\\
\textbf{rephrasing\_approach}: ["Keep the expressions for y1 and y2 intact", "Request the determination of the relationship between x1*x2 and y1*y2", "Specify the possible relationships (>, <, =)"]\\
\textbf{problem\_integrity\_preservation}: ["Maintain all original expressions and conditions", "Remove comparative aspect entirely"]\\
\textbf{answer\_format\_specification}: ["Answer should be expressed as one of the following: >, <, or =", "Include the relationship between x1*x2 and y1*y2 in the answer"]\\
\textbf{is\_multiple\_choice}: true\\
    \textcolor{blue}{\textbf{Reformulation:}} Given distinct positive real numbers $x_1, x_2$ and a real number $a$ in the interval (0,1), define $y_1 = x_1/(1+a) + a*x_2/(1+a)$ and $y_2 = x_2/(1+a) + a*x_1/(1+a)$. Determine the relationship between $x_1*x_2$ and $y_1*y_2$. Express your answer as one of the following: >, <, or =.\\
    \textcolor{blue}{\textbf{Judgement:}} The reformulated problem still implies a multiple-choice format with the options >, <, or =, which may limit the answer choices and does not fully open the problem to exploration.\\
    \end{text_full}
    \caption{\textbf{Examples of the reformulation process} starting from the original problem, we extract key information, and then use that information to reformulate the problem as an open-ended question. Finally, a critical evaluation of the reformulation determines whether the reformulation succeeded. We include an example of both a successful reformulation (top) and failed reformulation (bottom).}
    \label{fig:reformulation_examples}
\end{figure}

\clearpage

\section{Prompts}
\label{app:prompts}

\subsection{Llama-3.1-70B Filter prompts}
\label{app:filter_prompts}



\begin{figure}[h]
    \centering
    \begin{text_full}{Multiple Choice}
Given this question: \{problem\}\\
Is this a multiple choice question (a question that provides specific options to choose from, typically labeled as A, B, C, D or 1, 2, 3, 4)?
Return only "yes" or "no" without any additional explanation.
    \end{text_full}
\end{figure}

\begin{figure}[h]
    \centering
    \begin{text_full}{Proof}

Given this question: \{problem\} \\
Is this a mathematical proof question (a question that asks to prove a statement, theorem, or property...)?
Examples of proof indicators:
\\
- "Prove that..."
\\
- "Show that..."
\\
- "Demonstrate why..."
\\
- "Justify your answer..."
\\
- "Explain why..."
Here are examples of proof questions: \\
Example 1:  Given positive integers \(a\) and \(b\) such that \(b > a > 1\), and \(a\) does not divide \(b\), and a given sequence of positive integers \(\{b_n\}_{n=1}^{\infty}\) satisfying \(b_{n+1} \geq 2b_n\) for all positive integers \(n\). Does there always exist a sequence of positive integers \(\{a_n\}_{n=1}^{\infty}\) such that for all positive integers \(n\), \(a_{n+1} - a_n \in \{a, b\}\), and for all positive integers \(m\) and \(l\) (which can be the same), \(a_m + a_l \notin \{b_n\}\) for all \(n\)? \\
Example 2: Let \( f(x) = x^n, x \in D, n \in \mathbf{N}^{+} \). Determine whether \( f(x) \) is a solution to the functional inequality
\[ 
f(x) + f(1-x) > 1 
\]
If so, find the domain \( D \); if not, provide an explanation.
\\
Example 3: In a right angled-triangle $ABC$, $\angle{ACB} = 90^o$. Its incircle $O$ meets $BC$, $AC$, $AB$ at $D$,$E$,$F$ respectively. $AD$ cuts $O$ at $P$. If $\angle{BPC} = 90^o$, prove $AE + AP = PD$.
\\
Example 4: A(x,y), B(x,y), and C(x,y) are three homogeneous real-coefficient polynomials of x and y with degree 2, 3, and 4 respectively. we know that there is a real-coefficient polinimial R(x,y) such that $B(x,y)^2-4A(x,y)C(x,y)=-R(x,y)^2$. Show that there exist 2 polynomials F(x,y,z) and G(x,y,z) such that $F(x,y,z)^2+G(x,y,z)^2=A(x,y)z^2+B(x,y)z+C(x,y)$ if for any x, y, z real numbers $A(x,y)z^2+B(x,y)z+C(x,y)\ge 0$
\\
Example 5: Prove \[\frac{1}{\cos 0^\circ \cos 1^\circ} + \frac{1}{\cos 1^\circ \cos 2^\circ} + \cdots + \frac{1}{\cos 88^\circ \cos 89^\circ} = \frac{\cos 1^\circ}{\sin^2 1^\circ}.\]

Here are examples of non-proof questions:\\
Example 1: In a $100 \times 25$ rectangular table, each cell is filled with a non-negative real number. The number in the $i$-th row and $j$-th column is denoted by $x_{i, j}$ $(i=1,2,\ldots, 100; j=1,2,\ldots, 25)$ (Table 1). The numbers in each column of Table 1 are then reordered in descending order to create Table 2 such that $x_{1, j}^{\prime} \geq x_{2, j}^{\prime} \geq \cdots \geq x_{100, j}^{\prime}$ $(j=1,2,\ldots, 25)$. Find the smallest natural number $k$ such that if the numbers in Table 1 satisfy $\sum_{j=1}^{25} x_{i, j} \leq 1$ $(i=1,2,\ldots, 100)$, then for $i \geq k$, Table 2 satisfies $\sum_{j=1}^{25} x_{i, j}^{\prime} \leq 1$ $(i=1,2,\ldots, 100)$.

Example 2: We are given $2n$ natural numbers
\[1, 1, 2, 2, 3, 3, \ldots, n - 1, n - 1, n, n.\]
Find all $n$ for which these numbers can be arranged in a row such that for each $k \leq n$, there are exactly $k$ numbers between the two numbers $k$.

Example 3: Determine all positive integers $n$, $n\ge2$, such that the following statement is true: If $(a_1,a_2,...,a_n)$ is a sequence of positive integers with $a_1+a_2+\cdots+a_n=2n-1$, then there is block of (at least two) consecutive terms in the sequence with their (arithmetic) mean being an integer.

Example 4: Turbo the snail sits on a point on a circle with circumference $1$. Given an infinite sequence of positive real numbers $c_1, c_2, c_3, \dots$, Turbo successively crawls distances $c_1, c_2, c_3, \dots$ around the circle, each time choosing to crawl either clockwise or counterclockwise.
Determine the largest constant $C > 0$ with the following property: for every sequence of positive real numbers $c_1, c_2, c_3, \dots$ with $c_i < C$ for all $i$, Turbo can (after studying the sequence) ensure that there is some point on the circle that it will never visit or crawl across.

Example 5: For an even integer positive integer $n$ Kevin has a tape of length $4 n$ with marks at $-2 n,-2 n+1, \ldots, 2 n-1,2 n$. He then randomly picks $n$ points in the set $-n,-n+1,-n+2, \ldots, n-1, n$, and places a stone on each of these points. We call a stone 'stuck' if it is on $2 n$ or $-2 n$, or either all the points to the right, or all the points to the left, all contain stones. Then, every minute, Kevin shifts the unstuck stones in the following manner: He picks an unstuck stone uniformly at random and then flips a fair coin. If the coin came up heads, he then moves that stone and every stone in the largest contiguous set containing that stone one point to the left. If the coin came up tails, he moves every stone in that set one point right instead. He repeats until all the stones are stuck. Let $p_{k}$ be the probability that at the end of the process there are exactly $k$ stones in the right half. Evaluate $\frac{p_{n-1}-p_{n-2}+p_{n-3}-\ldots+p_{3}-p_{2}+p_{1}}{p_{n-1}+p_{n-2}+p_{n-3}+\ldots+p_{3}+p_{2}+p_{1}}$ in terms of $n$.

Example 6: A 0-1 sequence of length $2^k$ is given. Alice can pick a member from the sequence, and reveal it (its place and its value) to Bob. Find the largest number $s$ for which Bob can always pick $s$ members of the sequence, and guess all their values correctly.
Alice and Bob can discuss a strategy before the game with the aim of maximizing the number of correct guesses of Bob. The only information Bob has is the length of the sequence and the member of the sequence picked by Alice.
\\
Return only "yes" or "no" without any additional explanation.
    \end{text_full}
\end{figure}

\begin{figure}[h]
    \centering
    \begin{text_full}{Yes/No}
Given this question: \{problem\}
Is this a yes/no question (a question that asks to choose between two options, typically labeled as yes or no)?
Return only "yes" or "no" without any additional explanation.
    \end{text_full}
\end{figure}

\begin{figure}[h]
    \centering
    \begin{text_full}{True/False}
Given this question: \{problem\}
Is this a true/false question (a question that asks to choose between two options, typically labeled as true or false)?
Return only "true" or "false" without any additional explanation.
    \end{text_full}
\end{figure}

\begin{figure}[h]
    \centering
    \begin{text_full}{Multiple Part}
Your task is to determine if the given question contains multiple sub-questions, sub-parts, or sub-tasks.
A multi-part question requires separate answers for different components, rather than a single comprehensive answer.
Besides that, if the question is multiple choice and only requires to select one option, it is not a multi-part question. \\
Here are examples of multi-part questions that require multiple distinct answers: \\
Example 1: Given the set $M=\{0,1\}$, $A=\{(x,y)|x\in M, y\in M\}$, $B=\{(x,y)|y=-x+1\}$. 1. Please list the elements of set $A$. 2. Find $A\cap B$ and list all subsets of $A\cap B$. \\
Example 2: In the Cartesian coordinate system $xOy$, the parametric equation of curve $C_1$ is $\begin{cases} x=\cos \theta \\ y=1+\sin \theta \end{cases}$ (where $\theta$ is the parameter), and the equation of curve $C_2$ is $\frac{x^2}{1}+\frac{y^2}{2}=1$. With $O$ as the pole and the non-negative half-axis of $x$ as the polar axis, a polar coordinate system is established with the same unit of length as the Cartesian coordinate system $xOy$. (1) Find the polar equations of curves $C_1$ and $C_2$; (2) The ray $\theta =\frac{\pi }{3}(\rho > 0)$ intersects curve $C_1$ at point $A$ (other than the pole) and intersects curve $C_2$ at point $B$. Find $|AB|$.\\
Example 3: Given the function $f(x)=|x+2|-|2x-a|$, $(a\in\mathbb{R})$. (I) When $a=3$, solve the inequality $f(x) > 0$; (II) When $x \in [0, +\infty)$, $f(x) < 3$ always holds, find the range of $a$.\\
Example 4: Given an ellipse $C_1$: $\frac{x^2}{a^2} + \frac{y^2}{b^2} = 1$ ($a > b > 0$) with a major axis length of 4 and an eccentricity of $\frac{1}{2}$, where $F_1$ and $F_2$ are its left and right foci, respectively. A moving circle passes through point $F_2$ and is tangent to the line $x = -1$. (I) (i) Find the equation of the ellipse $C_1$; (ii) Find the equation of the trajectory of the center $C$ of the moving circle; (II) On the curve $C$, there are two points $M$ and $N$, and on the ellipse $C_1$, there are two points $P$ and $Q$, satisfying that $MF_2$ and $\overrightarrow{NF_2}$ are collinear, $\overrightarrow{PF_2}$ and $\overrightarrow{QF_2}$ are collinear, and $\overrightarrow{PF_2} \cdot \overrightarrow{MF_2} = 0$, find the minimum value of the area of quadrilateral $PMQN$. \\
Example 5: In the rectangular coordinate system $xOy$, a polar coordinate system is established with the coordinate origin as the pole and the positive semi-axis of the $x$-axis as the polar axis. The polar coordinate equation of circle $C$ is $\rho^2 - 2m\rho\cos\theta + 4\rho\sin\theta = 1 - 2m$. (1) Find the rectangular coordinate equation of $C$ and its radius. (2) When the radius of $C$ is the smallest, the curve $y = \sqrt{3}|x - 1| - 2$ intersects $C$ at points $A$ and $B$, and point $M(1, -4)$. Find the area of $\triangle MAB$. \\
Here are examples of single-part questions that require only one answer: \\
Example 1: Ancient Greek mathematicians from the Pythagorean school studied various polygonal numbers, such as triangular numbers 1, 3, 6, 10, ..., with the $n$-th triangular number being $\frac{n(n+1)}{2} = \frac{1}{2}n^{2} + \frac{1}{2}n$. Let the $n$-th $k$-sided polygon number be denoted as $N(n, k)$ ($k \geq 3$). Below are the expressions for the $n$-th number of some $k$-sided polygon numbers:  \\Triangular numbers $N(n, 3) = \frac{1}{2}n^{2} + \frac{1}{2}n$  \\Square numbers $N(n, 4) = n^{2}$  \\Pentagonal numbers $N(n, 5) = \frac{3}{2}n^{2} - \frac{1}{2}n$  \\Hexagonal numbers $N(n, 6) = 2n^{2} - n$  \\...  \\From this, we can deduce the expression for $N(n, k)$ and calculate $N(8, 12) = \_\_\_\_\_\_$.' \\
Example 2: Let x be the dividend, y be the divisor, z be the quotient, and r be the remainder. If $y = 3(y_1 + y_2) + 4, z = 2z_1^2 - z_2, r = 3r_1 + 2,$ and $x = 2x_1 y_1 - x_2 + 10,$ find the values of $x, y, z$, and $r$, given that $y_1 = 2, y_2 = 3, z_1 = 3, z_2 = 5, r_1 = 1, x_1 = 4,$ and $x2 = 6$. \\
Example 3: If  $x_{1}, x_{2},\ldots ,x_{n}$  are positive real numbers with  $x_{1}^2+x_2^{2}+\ldots +x_{n}^{2}=1$ , find the minimum value of  $\sum_{i=1}^{n}\frac{x_{i}^{5}}{x_{1}+x_{2}+\ldots +x_{n}-x_{i}}$ . \\
Example 4: Given that the value of the function \( f(x) = \frac{1}{(x-2)^{2}} - 2x + \cos 2\theta - 3 \sin \theta + 2 \) is always positive for \( x \in (-\infty, 2) \), determine the range of the parameter \( \theta \) within the interval \( (0, \pi) \). \\
Example 5: A transparent, sealed cubic container is exactly half filled with water. When this cube is rotated arbitrarily, the shape of the water surface inside the container can be: (1) triangle; (2) rectangle; (3) square; (4) regular hexagon. Among these, the correct conclusions are \_\_\_\_\_\_\_\_\_\_.

    \end{text_full}
\end{figure}

\begin{figure}[h]
    \centering
    \begin{text_full}{Llama-3.1-8B Rollout Prompt}
You are a math expert. Given the following math problem, provide your solution in Latex format. Always format your final answer in perfect LaTeX \textbackslash boxed\{final\_answer\} format.
    \end{text_full}
    \label{app:llama8b_rollout}
\end{figure}

\begin{figure}[h]
    \centering
    \begin{text_full}{Llama-3.1-405B Rollout Prompt (part 1)}
From this moment forward, adopt the persona of a tenured mathematics professor. You receive math problems and you solve them, step by step, following the formatting instructions below.\\

\# Mathematical Solution Formatting Guidelines\\

\#\# Mathematical Notation Rules\\

**Use \$\$ \$\$ for:**\\
- Standalone equations\\
- Complex mathematical expressions\\
- Multi-line equations\\
- Final answers\\

**Use single \$ \$ for:**\\
- Simple inline expressions\\
- Variables\\
- Numbers with mathematical meaning\\
- Parts of expressions being discussed\\

**For equations with multiple lines:**\\
`latex\\
\$\$\textbackslash\textbackslash begin\{aligned\}\\
equation1 \&= expression1 \textbackslash\textbackslash\textbackslash\textbackslash \\
equation2 \&= expression2\\
\textbackslash\textbackslash end\{aligned\}\$\$\\
`\\

**Place the final answer inside \textbackslash boxed\{answer\_here\}**\\
After solving and verifying your full solution, write the final answer in \textbackslash\textbackslash boxed\{answer\} notation.\\

\#\# Thought Structure\\
Each thought must:\\
1. Begin with active voice ("I" statements or direct observations)\\
2. Express exactly ONE logical step\\
3. Integrate mathematical notation naturally within sentences\\
4. Use appropriate notation based on context\\
5. No more than 1 sentence + interleaved math notation long\\

\#\#\# Example:
"I can see that when $x = 2$, the expression $y = x^2 + 3x + 1$ becomes $y = 4 + 6 + 1 = 11$."\\

\#\# Common Mathematical Patterns\\
- Use \$\$ \$\$ for equations being solved\\
- Use \$ \$ for discussing components: "where \$m\$ is the slope"\\
- Keep expressions inline when discussing specific parts\\
- Use block notation (\$\$) for key steps and results\\

\#\# Visual Formatting\\
- Two blank lines between thoughts (\textbackslash n \textbackslash n)\\
- No bullet points or numbered lists\\
- No section headers within the solution\\
- Mathematical expressions should flow naturally within sentences\\
- No explicit step labeling\\
- Human-style reasoning\\

---\\
    \end{text_full}
    \label{app:llama405b_rollout}
\end{figure}

\begin{figure}
    \centering
    \begin{text_full}{Llama-3.1-405B Rollout Prompt (part 2)}
\#\# Examples\\

\#\#\# Problem 1:\\

A standard die is rolled six times. What is the probability that the product of all six rolls is odd? Express your answer as a common fraction.\\

\#\#\# Solution 1:\\

I need to find the probability by dividing favorable outcomes by total outcomes.\\

I know that the total number of possible outcomes when rolling a die six times is $6^6$.\\

For the product to be odd, each individual roll must be odd.\\

Looking at a standard die, the odd numbers are 1, 3, and 5.\\

Therefore, for each roll, I have only 3 choices to maintain an odd product.\\

The total number of favorable outcomes is $3^6$ since I have 3 choices for each of the 6 rolls.\\

The probability is thus $\frac{3^6}{6^6}$.

This simplifies to $\left( \frac{1}{2} \right)^6 =$ \textbackslash\textbackslash boxed$\{\frac{1}{64}\}$.\\

\#\#\# Problem 2:\\

Solve over the integers: $2^a+4^b+8^c=328$

\#\#\# Solution 2:\\

I notice that all terms are powers of 2, so I can rewrite everything with base 2.

I know that $4 = 2^2$ and $8 = 2^3$, so the equation becomes $2^a + 2^{2b} + 2^{3c} = 328$.

To solve this, I can decompose 328 into powers of 2.

The largest power of 2 less than 328 is 256 ($2^8$).

Subtracting 256 from 328 leaves 72.

The largest power of 2 less than 72 is 64 ($2^6$).

Subtracting 64 leaves 8 ($2^3$).

Therefore, $328 = 2^8 + 2^6 + 2^3$.
    \end{text_full}
\end{figure}

\begin{figure}
    \centering
    \begin{text_full}{Llama-3.1-405B Rollout Prompt (part 3)}
Comparing terms: $2^a = 2^8$, $2^{2b} = 2^6$, and $2^{3c} = 2^3$.

Solving these equations: $a = 8$, $b = 3$, and $c = 1$.

My final answer is \textbackslash\textbackslash boxed\{(a, b, c) = (8, 3, 1)\}\\

\#\#\# Problem 3:\\

Find the equation of the circle which passes through $(2, 3)$ and $(4, 5)$ and whose center lies on the straight line $y - 4x + 3 = 0$.\\

\#\#\# Solution 3:\\

I know that the center lies on the line, so I can write its coordinates as $(x, 4x - 3)$.

The radius can be found using the distance from the center to $(2, 3)$.

Using the distance formula, I get $r^2 = (x - 2)^2 + (4x - 3 - 3)^2 = (x - 2)^2 + (4x - 6)^2$.

Expanding and simplifying: $r^2 = 17x^2 - 52x + 40$.

Since the circle also passes through $(4, 5)$, I can write $r^2 = (x - 4)^2 + (4x - 3 - 5)^2$.

This simplifies to $r^2 = 17x^2 - 72x + 80$.

Since both expressions equal $r^2$, I can write $17x^2 - 52x + 40 = 17x^2 - 72x + 80$.
    \end{text_full}
\end{figure}

\begin{figure}
    \centering
    \begin{text_full}{Llama-3.1-405B Rollout Prompt (part 4)}
I can now simplify: $20x = 40$, so $x = 2$.

The center is therefore $(2, 4(2) - 3) = (2, 5)$.

The radius squared is $r^2 = 17(2)^2 - 52(2) + 40 = 16$, so $r = 4$.

The equation of the circle is $(x - 2)^2 + (y - 5)^2 = 16$.

The final answer is the full equation of the circle, so the correct final answer is \textbackslash\textbackslash boxed\{$(x - 2)^2 + (y - 5)^2 = 16$\}
    \end{text_full}
\end{figure}

\subsection{Reformulation Prompts}
\label{app:reformulation_prompts}

\begin{figure}
    \centering
    \begin{text_full}{Multiple Choice Reformulation (Part 1)}

You are an AI assistant specializing in transforming multiple-choice math problems into open-ended, solvable questions suitable for an automatic grading system that relies on regex pattern matching within `\boxed{}`. Your task is to reformulate the given problem while adhering to specific guidelines.\\

**Rules:**\\
- Never turn problems into proofs, all problems should have an answer that is machine-verifiable\\
- Reformulated problems should have the same final answer as the original problem (just not in the multiple-choice format)\\
- It must be reasonably clear from the problem what the student is expected to write inside of \boxed{final\_answer}. E.g. "What are its key properties and characteristics?" is almost always a bad reformulation because it is ambiguous what the student should write inside of the `\boxed{}`. "Express your answer in centimeters." is a good answer format specification because it does NOT give away the answer but does specify a clear format.\\
- Do NOT reformulate problems that are not actually multiple choice or are impossible to reformulate\\
- If a problem is not multiple choice, return "N/A" for the reformulated problem and other "N/A" fields\\

**Follow these steps:**\\

1. **Analyze the original problem:**\\
    - Identify the core mathematical concept.\\
    - Note any crucial information, including numerical values, equations, and key terms.\\
    - Determine the structure of the problem (e.g., scenario-based, direct question, multi-step).\\
    - Preserve all mathematical notations, symbols, and formatting as they appear in the original problem.\\

2. **Remove multiple-choice options:**\\
    - Eliminate all answer choices and their labels (A, B, C, etc.).\\
    - If any options contain information essential to solving the problem, incorporate that information into the main problem statement.\\

3. **Rephrase the question:**\\
    - Transform the question into an open-ended format that requires a direct numerical or algebraic answer.\\
    - Ensure the rephrased question is clear, unambiguous, and uses language appropriate for the student's level.\\

4. **Maintain problem integrity:**\\
    - Keep all original numerical values, equations, figures, and key terms intact.\\
    - Preserve any scenarios, dialogues, or conditional information crucial for solving the problem.\\
    - Do not introduce new information, alter units of measurement, or change the mathematical intent of the problem.\\
    - If the problem references diagrams or figures, ensure that any necessary descriptions are included.\\

5. **Specify the answer format:**\\
    - Instruct the student to provide their answer using the `\boxed{}` format.\\
    - Do not include placeholders like `\boxed{N}`; instead, guide the student to input their calculated answer within the boxed format.\\
    - For example, "Provide your answer in the form `\boxed{\text{your answer here}}`."\\

6. **Final check:**\\
    - Ensure the reformulated problem contains all necessary information for independent solving.\\
    - Verify that the problem hasn't become easier or harder than the original.\\
    - Check for any common errors, such as unit inconsistencies or typographical mistakes.\\
    - Confirm that no hints, solution methods, or additional explanations have been inadvertently added.\\

7. **Is actually multiple choice:**\\
    - Some problems are not actually multiple choice and do NOT actually need to be reformulated.\\
    - If the problem is **NOT** multiple choice, do NOT reformulate it! Note here if it is not multiple choice and return "N/A" for the reformulated problem.\\

**Before providing the final reformulated problem, please create a `reformulation\_process` dictionary (for internal use; do not include this dictionary or its content in the final problem). The dictionary should have the following exact keys:**\\

- `"core\_mathematical\_concept"`: Summarize the core mathematical concept.\\
- `"key\_information\_extraction"`: List key information (numerical values, equations, terms).\\
- `"problem\_structure\_analysis"`: Identify the problem structure.\\
- `"multiple\_choice\_removal\_strategy"`: Plan how to remove multiple-choice options.\\
- `"rephrasing\_approach"`: Outline the rephrasing strategy.\\
    \end{text_full}
\end{figure}

\begin{figure}
    \centering
    \begin{text_full}{Multiple Choice Reformulation (Part 2)}
    - `"problem\_integrity\_preservation"`: Note how to maintain problem integrity.\\
    - `"answer\_format\_specification"`: Specify the answer format.\\
    - `"is\_multiple\_choice"`: Whether the problem is actually multiple choice.\\
    
**This structured approach will help ensure accuracy and completeness in your final output.**\\

**After your reformulation process, present the reformulated problem as plain text, making sure it meets all the criteria outlined above. Do not include the `<reformulation\_process>` tags or any of their content in the final problem presented to the student.**\\

**Examples:**\\
```\\
<example>\\
<problem>\\
Three squares with the same center and corresponding parallel sides are drawn. The distance between the sides of successive squares is 3 units, and the side length of the largest square is 22 units. What is the perimeter of the smallest square?\\
(A) 40\\
(B) 64\\
(C) 100\\
(D) 10\\
(E) 20\\
</problem>\\

<reformulation\_process>\\
{\\
    "core\_mathematical\_concept": "Geometry - concentric squares and their dimensions",\\
    "key\_information\_extraction": [\\
        "Three concentric squares",\\
        "Distance between sides of successive squares: 3 units",\\
        "Side length of the largest square: 22 units",\\
        "Need to find the perimeter of the smallest square"\\
    ],\\
    "problem\_structure\_analysis": "Direct question based on given information",\\
    "multiple\_choice\_removal\_strategy": [\\
        "Eliminate all answer choices and their labels",\\
        "No essential information is lost by removing the options"\\
    ],\\
    "rephrasing\_approach": [\\
        "Keep the description of the squares intact",\\
        "Rephrase the question to ask for the perimeter in an open-ended format",\\
        "Specify units and number format requirements"\\
    ],\\
    "problem\_integrity\_preservation": [\\
        "Preserve all original numerical values and key terms",
        "Do not add any new information or hints"\\
    ],\\
    "answer\_format\_specification": [\\
        "Answer should be expressed in units",\\
        "Answer should be given as a whole number",\\
        "No decimal places required based on original options"\\
    ],\\
    "is\_multiple\_choice": true\\
}\\
</reformulation\_process>\\

<reasoning>\\
Let me think about this step by step:\\

1. Analyzing the multiple choice aspect:\\
   - The options are all whole numbers: 40, 64, 100, 10, 20\\
   - No units are specified in the options but units are used in the problem\\
   - The options suggest an exact answer is expected, not an approximation\\

2. Looking at the essential information:\\
   - The geometric relationships are clearly stated\\
   - All measurements are given in units\\
   - No complex transformations or unit conversions needed\\
    \end{text_full}
\end{figure}

\begin{figure}
    \centering
    \begin{text_full}{Multiple Choice Reformulation (Part 3)}
3. What needs to be preserved:\\
   - The concentric squares description\\
   - The 3-unit distance between squares\\
   - The 22-unit largest square side length\\
   - The focus on perimeter rather than area\\

4. What needs to be clarified:\\
   - The answer should include units\\
   - A whole number is expected\\
   - No decimal places needed\\

5. How to make it open-ended:\\
   - Remove all multiple choice options\\
   - Keep the geometric scenario exactly as is\\
   - Add specific answer format requirements\\
   - Maintain the same level of difficulty\\

6. Final considerations:\\
   - The reformulated question should be solvable without reference to the original options\\
   - The specifications should ensure an unambiguous answer format\\
   - The mathematical challenge should remain unchanged\\
</reasoning>\\

<reformulated\_problem>\\
Three squares with the same center and corresponding parallel sides are drawn. The distance between the sides of each successive square is 3 units, and the side length of the largest square is 22 units. Find the perimeter of the smallest square. Express your answer in units as a whole number.\\
</reformulated\_problem>\\
</example>\\
```\\

```\\
<example>\\
<problem>\\
The teacher picked 9 out of 13 cards labeled with numbers 1 to 13 and stuck them on the foreheads of 9 students. Each student can see the numbers on the other 8 students' foreheads but not their own. The teacher asked, "Those who know the number of divisors of their own number, please raise your hand." Two students raised their hands. After they put their hands down, three students had the following conversation.\\
- A: "I know my number now."\\
- B: "I still don't know what my number is, but I have figured out whether it is odd or even."\\
- C: "My number is 2 less than B's number and 1 more than A's number."\\
What is the sum of the numbers on the four cards that were not picked?\\
</problem>\\

<reformulation\_process>\\
{\\
"core\_mathematical\_concept": "Number theory and logical deduction with combinatorics",\\
"key\_information\_extraction": [\\
    "13 cards numbered from 1 to 13",\\
    "9 cards picked and assigned to students",\\
    "Students can see others' numbers but not their own",\\
    "Two students know their number based on number of divisors",\\
    "Conversation between students A, B, and C revealing relationships",\\
    "Need to find sum of the 4 unpicked cards"\\
],\\
"problem\_structure\_analysis": "Complex logical deduction problem combining number theory with information from multiple stages",\\
"multiple\_choice\_removal\_strategy": "Not applicable as this is not a multiple choice question",\\
"rephrasing\_approach": "Not needed as problem is already in appropriate format",\\
"problem\_integrity\_preservation": "No modifications needed",\\
"answer\_format\_specification": "Answer should be expressed as a single integer (sum of the unpicked cards)",\\
"is\_multiple\_choice": false\\
}\\
</reformulation\_process>\\
    \end{text_full}
\end{figure}

\begin{figure}
    \centering
    \begin{text_full}{Multiple Choice Reformulation (Part 4)}
<reasoning>\\
Let me think about this step by step:\\

1. First, assess if reformulation is needed:\\
   - This is not a multiple choice question\\
   - It already asks for a specific numerical answer\\
   - The format is clear and unambiguous\\

2. Analyzing the problem structure:\\
   - It's a complex logic puzzle\\
   - All information provided is necessary\\
   - The sequence of events matters\\
   - The dialogue provides crucial information\\

3. Consider what might need clarification:\\
   - Answer format is implicitly clear (a sum)\\
   - No unit specification needed\\
   - No decimal places involved (working with integers 1-13)\\
   - No ambiguity in what constitutes an answer\\

4. Check for potential improvements:\\
   - Current format is optimal for the problem type\\
   - Adding structure might make it harder to follow\\
   - The narrative flow is important for understanding\\

5. Final decision:\\
   - No reformulation needed\\
   - Problem is already in ideal open-ended format\\
   - Answer requirements are clear (single integer sum)\\
</reasoning>\\

<reformulated\_problem>\\
N/A\\
</reformulated\_problem>\\
</example>\\
```\\

```\\
<example>\\
<problem>\\
In the cube shown, $L$ and $K$ are midpoints of adjacent edges $AD$ and $AB$, respectively. The perpendicular distance from vertex $F$ to the line segment $LK$ is 10 units. Which of the following measurements is closest to the volume of the cube after subtracting 3 units?\\

(A) 200 in$^3$\\
(B) 500 cm$^3$\\
(C) 0.01039 m$^3$\\
(D) Tie\\
</problem>\\

<reformulation\_process>\\
\{\\
    "core\_mathematical\_concept": "3D geometry involving cube volume calculation",\\

    "key\_information\_extraction": [\\
        "Points L and K are midpoints of adjacent edges AD and AB",\\
        "Perpendicular distance from F to LK is 10 units",\\
        "Need to subtract 3 units from final volume"\\
    ],\\

    "problem\_structure\_analysis": "Geometry problem requiring volume calculation",\\

    "multiple\_choice\_removal\_strategy": [\\
        "Remove all comparisons",\\
        "Ask for direct volume calculation"\\
    ],
    \end{text_full}
\end{figure}

\begin{figure}
    \centering
    \begin{text_full}{Multiple Choice Reformulation (Part 5)}
    "rephrasing\_approach": [\\
        "Keep geometric relationships intact",\\
        "Request exact volume calculation",\\
        "Specify units and precision"\\
    ],\\

    "problem\_integrity\_preservation": [\\
        "Maintain all numerical values and geometric relationships",\\
        "Remove comparative aspect entirely"\\
    ],\\

    "answer\_format\_specification": [\\
        "Answer should be expressed in cubic meters (m³)",\\
        "Round to 5 decimal places",\\
        "Include units in answer"\\
    ],\\

    "is\_multiple\_choice": true\\
\}\\
</reformulation\_process>\\
<reasoning>\\
Let me think about this step by step:\\

1. First, I need to identify what makes this multiple choice:\\

   - The core task is actually calculating a specific volume\\
   - The multiple choice aspect is just about comparing to different unit measurements\\
   - The comparison isn't fundamental to the mathematical concept being tested\\

2. Looking at the original answer choices:\\

   - They use different units (in³, cm³, m³)\\
   - The m³ value has 5 decimal places (0.01039)\\
   - This suggests m³ with 5 decimal places is an appropriate precision level\\

3. How to remove the multiple choice aspect:\\

   - Instead of asking which measurement is closest\\
   - Ask for the exact calculation\\
   - Need to specify one consistent unit to avoid ambiguity\\

4. What to preserve:\\

   - All geometric relationships\\
   - The 3-unit subtraction\\
   - The original scenario and values\\

5. What needs to be added:\\

   - Specific unit requirement (m³)\\
   - Decimal place requirement (5 places)\\
   - Clear instruction about including units\\

6. The reformulated question should:\\

   - Be completely open-ended\\
   - Have a unique, calculable answer\\
   - Remove all traces of comparison\\
   - Maintain the same difficulty level\\

</reasoning>\\
<reformulated\_problem>\\
In the cube shown, \( L \) and \( K \) are midpoints of adjacent edges \( AD \) and \( AB \), respectively. The perpendicular distance from vertex \( F \) to the line segment \( LK \) is 10 units. Calculate the volume of the cube after subtracting 3 cubic units. Express your answer in cubic meters (m³), rounded to 5 decimal places.\\
</reformulated\_problem>\\
</example>\\
```
    \end{text_full}
\end{figure}

\begin{figure}
    \centering
    \begin{text_full}{Multiple Choice Reformulation Judge (Part 1)}
You are a IMO judge, specialized in evaluating the quality of IMO math problem reformulations. Your task is to critically analyze whether a reformulated problem truly meets the criteria for being open-ended, challenging, and maintains the original problem's integrity.\\

**Important:**\\
- Problems you review are being transformed from multiple choice problems into open-ended, solvable questions suitable for an automatic grading system that relies on regex pattern matching within `\boxed{}`. It is VERY important that even though the problem is no longer multiple choice, it still has a single, definite answer.\\
- It must be reasonably clear from the problem what the student is expected to write inside of \boxed{final\_answer}. E.g. "What are its key properties and characteristics?" is almost always a bad reformulation because it is ambiguous what the student should write inside of the `\boxed{}`. "Express your answer in centimeters." is a good answer format specification because it does NOT give away the answer but does specify a clear format.\\

Key aspects to evaluate:\\

1. Hidden Multiple Choice\\
    - Check if the reformulation still effectively presents multiple choice options by:\\
        - Embedding a limited/fixed set of choices (e.g. 4 or 5 options) within the problem text that limits the answer choices to one of those options\\
        - Asking to compare with specific values\\
        - Limiting answers to specific options\\
    - Flag any reformulations that are just disguised multiple choice questions\\

Example:\\

<problem>\\
For real numbers $t \neq 0,$ the point \[(x,y) = \left( \frac{t + 1}{t}, \frac{t - 1}{t} \right)\]is plotted. All the plotted points lie on what kind of curve? (A) Line (B) Circle (C) Parabola (D) Ellipse (E) Hyperbola Enter the letter of the correct option.\\
</problem>\\

Reformulated failed example (all multiple choice aspect still present):\\
<bad\_reformulated\_problem>\\
For real numbers $t \neq 0,$ the point \[(x,y) = \left( \frac{t + 1}{t}, \frac{t - 1}{t} \right)\]is plotted. What type of curve do all the plotted points lie on? Provide your answer as a specific curve type (e.g., line, circle, parabola, ellipse, hyperbola).\\
</bad\_reformulated\_problem>\\

Reformulated successful example (multiple choice aspect removed):\\
<reformulated\_problem>\\
For real numbers \( t \neq 0 \), the point
\[
(x, y) = \left( \frac{t + 1}{t}, \frac{t - 1}{t} \right)
\]
is plotted. Determine the type of coordinate geometry curve on which all the plotted points lie.\\
</reformulated\_problem>\\

2. Mathematical Integrity\\
    - Verify that the mathematical difficulty remains unchanged\\
    - Ensure no accidental hints or simplifications were introduced\\
    - Check that all necessary information was preserved\\
    - Confirm no extraneous information was added\\

Example 1:\\
This problem was not actually multiple choice, but it does imply options that are clearly missing/omitted.\\
<problem>\\
A resident wants to renovate their house and buys several strips of wood, each with a length of 0.7 meters and 0.8 meters. By connecting some of these wood strips, many different lengths of wood can be obtained. For example, \(0.7 + 0.7 = 1.4\) meters, \(0.7 + 0.8 = 1.5\) meters, etc. From the options below, what length of wood strip cannot be obtained by connecting these wood strips?\\
</problem>
    \end{text_full}
\end{figure}

\begin{figure}
    \centering
    \begin{text_full}{Multiple Choice Reformulation Judge (Part 2)}
Example 2:\\
<problem>\\
$\triangle ABC$ is inscribed in a semicircle of radius $r$ so that its base $AB$ coincides with diameter $AB$. Point $C$ does not coincide with either $A$ or $B$. Let $s=AC+BC$. Then, for all permissible positions of $C$: $\textbf{(A)}\ s^2\le8r^2\qquad \textbf{(B)}\ s^2=8r^2 \qquad \textbf{(C)}\ s^2 \ge 8r^2 \qquad\\ \textbf{(D)}\ s^2\le4r^2 \qquad \textbf{(E)}\ s^2=4r^2$\\
</problem>\\

The reformulation is flawed because it prematurely focuses on a single inequality (\(s^2 \leq 8r^2\)) rather than inviting exploration of the entire range of \(s^2\). It biases the solver, limits generality, and reduces the problem's open-ended nature.\\
<bad\_reformulated\_problem>\\
$\triangle ABC$ is inscribed in a semicircle of radius $r$ so that its base $AB$ coincides with diameter $AB$. Point $C$ does not coincide with either $A$ or $B$. Let $s=AC+BC$. Prove or disprove the inequality $s^2\le8r^2$ for all permissible positions of $C$.\\
</bad\_reformulated\_problem>\\

An open-ended revision that requires the same analysis and leads to the same conclusion\\
<reformulated\_problem>\\
Let triangle \( ABC \) be inscribed in a semicircle of radius \( r \), with its base \( AB \) coinciding with the diameter \( AB \). Point \( C \) lies on the semicircle but does not coincide with \( A \) or \( B \). Let \( s = AC + BC \). Determine the maximum possible value of \( s^2 \) in terms of \( r \), and prove that \( s^2 \leq 8r^2 \) for all permissible positions of \( C \).\\
</reformulated\_problem>\\

Example 3:\\
<problem>\\
For real numbers $t,$ the point \[(x,y) = \left( \frac{1 - t^2}{1 + t^2}, \frac{2t}{1 + t^2} \right)\]is plotted. All the plotted points lie on what kind of curve? (A) Line (B) Circle (C) Parabola (D) Ellipse (E) Hyperbola Enter the letter of the correct option.\\
</problem>\\

Although it removes the multiple-choice format, it does not truly open the problem to exploration. The answer asks for a description, which introduces too much variability in responses
<bad\_reformulated\_problem>\\
For real numbers $t,$ the point \[(x,y) = \left( \frac{1 - t^2}{1 + t^2}, \frac{2t}{1 + t^2} \right)\]is plotted. What kind of curve do all the plotted points lie on? Provide a brief description of the curve.\\
</bad\_reformulated\_problem>\\

This reformulation is clear, concise, and ensures consistent answers while remaining open-ended for exploration.
<reformulated\_problem>\\
For real numbers \( t \), consider the point
\[
(x, y) = \left( \frac{1 - t^2}{1 + t^2},\ \frac{2t}{1 + t^2} \right).
\]

Determine the type of curve on which all such points \((x, y)\) lie. State your answer as the name of the curve.\\
</reformulated\_problem>\\

Example 4:\\

<problem>\\
Define * as an operation on ordered pairs of real numbers, such that $(a, b) *(c, d)=$ $(a c+b d, a d+b c)$. If $(a, b) *(x, y)=(a, b)$, then which of the following is $(x, y)$? (A) $(0,0)$. (B) $(0,1)$. (C) $(1,0)$. (D) $(1,1)$. (E) $(1,-1)$.\\
</problem>\\

Ambiguous about whether the equation holds for all \((a, b)\) or specific values, lacks context to ensure a unique solution, and removes guidance provided by the multiple-choice format. Becomes unbounded.\\
<bad\_reformulated\_problem>\\
Define * as an operation on ordered pairs of real numbers, such that $(a, b) *(c, d)=$ $(a c+b d, a d+b c)$. If $(a, b) *(x, y)=(a, b)$, then calculate the ordered pair $(x, y)$. Express your answer in the format $(x, y)$, including parentheses and a comma.\\
</bad\_reformulated\_problem>
    \end{text_full}
\end{figure}

\begin{figure}
    \centering
    \begin{text_full}{Multiple Choice Reformulation Judge (Part 3)}
Ideal because it clarifies the scope (applies to all \((a, b)\)), ensures a unique solution, and balances open-ended exploration with clear grading criteria\\
<reformulated\_problem>\\
Define \( * \) as an operation on ordered pairs of real numbers, such that
\[
(a, b) * (c, d) = (ac + bd, ad + bc).
\]
Find the ordered pair \((x, y)\) that satisfies
\[
(a, b) * (x, y) = (a, b)
\]
for **all** real numbers \(a\) and \(b\). Express your answer as \((x, y)\).\\
</reformulated\_problem>\\

3. Answer Format Clarity\\
    - Evaluate if the answer format specification is:\\
        - Clear and unambiguous\\
        - Appropriate for the mathematical concept\\
        - Not overly unbounded or restrictive in a way that creates an open-ended problem out of it\\
    - Check if unit/precision requirements make sense for the problem\\

4. Problem Independence\\
    - Verify the reformulated problem can stand alone\\
    - Ensure it doesn't rely on knowledge of the original options\\
    - Check that answer requirements aren't derived solely from original choices\\

Example:\\
<problem>\\
Which of the following is a root of the equation \( x^2 - x - 6 = 0 \)?
(A) \( -3 \) (B) \( -2 \) (C) \( 2 \) (D) \( 3 \) (E) \( 6 \)\\
</problem>\\

Relies on original options, failing to stand independently and limiting exploration.\\
<bad\_reformulated\_problem>\\
Find a root of the equation \( x^2 - x - 6 = 0 \). Your answer must be one of the following: \( -3, -2, 2, 3, 6 \).\\
</bad\_reformulated\_problem>\\

Ideal because it ensures the solver identifies all roots without being constrained by the original options.\\
<reformulated\_problem>\\
Solve the quadratic equation \( x^2 - x - 6 = 0 \) and find all real roots. Provide your answers in increasing order.\\
</reformulated\_problem>\\

Remember: A truly open-ended reformulation should allow for calculation and expression of the answer
without any reference to or knowledge of the original multiple choice options.
    \end{text_full}
\end{figure}


\end{document}